\documentclass[10pt,twocolumn,letterpaper]{article}
\usepackage[pagenumbers]{cvpr}

\usepackage{booktabs}
\usepackage{multirow}
\usepackage{placeins}

\usepackage{amsmath,amssymb}
\usepackage{xcolor}
\usepackage{graphicx}
\usepackage{tikz}
\usetikzlibrary{arrows.meta,positioning,fit,calc,decorations.pathreplacing}
\definecolor{motc}{RGB}{66,133,244}    %
\definecolor{txtc}{RGB}{52,168,83}     %
\definecolor{mskc}{RGB}{150,150,150}   %
\definecolor{trnc}{RGB}{251,188,5}     %
\definecolor{t5c}{RGB}{171,71,188}     %
\tikzset{
  frame/.style={draw=motc!80!black, fill=motc!25, minimum width=0.42cm, minimum height=0.55cm, inner sep=0pt, font=\scriptsize},
  unit/.style={draw=txtc!80!black, fill=txtc!20, minimum height=0.55cm, inner sep=2pt, font=\scriptsize\sffamily},
  masked/.style={draw=mskc!90!black, fill=mskc!35, minimum width=0.42cm, minimum height=0.55cm, inner sep=1pt, font=\tiny\sffamily},
  block/.style={draw=black!70, fill=trnc!30, rounded corners=2pt, minimum height=0.7cm, font=\small\sffamily, align=center},
  t5block/.style={draw=black!70, fill=t5c!20, rounded corners=2pt, minimum height=0.7cm, font=\small\sffamily, align=center},
  lab/.style={font=\scriptsize\sffamily, align=center},
  arr/.style={-{Stealth[length=4pt]}, thick, black!70},
}
\newcommand{\flame}[1]{%
  \begin{scope}[shift={(#1)}, scale=0.16]
    \fill[red!80!orange] (0,0) .. controls (-1.1,0.4) and (-0.9,1.4) .. (0,2.3) .. controls (0.2,1.4) and (1.0,1.2) .. (0.6,0.5) .. controls (0.9,0.9) and (1.1,0.3) .. (0.7,0) -- cycle;
    \fill[yellow!90!orange] (0.15,0.15) .. controls (-0.45,0.4) and (-0.3,0.95) .. (0.1,1.35) .. controls (0.25,0.9) and (0.55,0.7) .. (0.45,0.15) -- cycle;
  \end{scope}}

\makeatletter
\renewcommand\paragraph{\@startsection{paragraph}{4}{\z@}{0pt}{-1em}{\normalfont\normalsize\bfseries}}
\makeatother

\definecolor{cvprblue}{rgb}{0.21,0.49,0.74}
\usepackage[pagebackref,breaklinks,colorlinks,allcolors=cvprblue]{hyperref}

\def\confName{CVPR}
\def\confYear{2027}

\title{MoGloss: Joint Motion and Pseudo-Gloss Pretraining for American Sign Language Understanding}
\author{Rui Hong \qquad Jana Ko\v{s}eck\'a\\
George Mason University\\
{\small Project page: \url{https://hongrui16.github.io/MoGloss}}
}

\begin{document}
\maketitle
\begin{abstract}

Sign language translation and recognition have been shown to benefit from various pretraining approaches. We propose a novel two-stage pretraining approach for the motion encoder, which first
learns to reconstruct masked motion, followed by joint masked prediction of pseudo-gloss units and motion. Pseudo-glosses serve as an intermediate structured representation of natural language text and are obtained from the captions by a large language model. We curate a corpus of about 1,100 hours of American Sign Language from five public datasets, estimate 2D pose for all clips and define an evaluation protocol. We then reproduce state-of-the-art translation methods and evaluate them using the same train and test splits.
We propose a translation model built on our pretraining strategy, and this model outperforms the previous state-of-the-art methods Uni-Sign and SHuBERT, as well as the YouTube-ASL baseline, on the How2Sign, ASL STEM Wiki and OpenASL datasets. Furthermore, the gain from the joint pretraining using motion and pseudo-gloss grows as translation data shrinks.
A classifier built on the same encoder also achieves state-of-the-art accuracy on ASL Citizen isolated sign recognition and outperforms our reproduction of Uni-Sign on WLASL2000.
\end{abstract}

\begin{figure}[t]
\centering
\includegraphics[width=\linewidth]{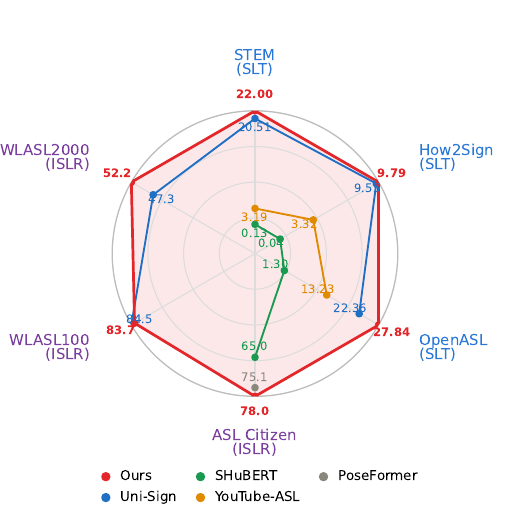}
\vspace{-14pt}
\caption{Results on SLT (BLEU-4, setting A) and ISLR (top-1 accuracy). Every method was trained and tested under the same protocol by us, except the published SHuBERT and PoseFormer results on ASL Citizen. Each axis is scaled to its best result, and a line is left open where a method has no result. Uni-Sign is the variant with RGB on translation and our pose-based reproduction on WLASL, and SHuBERT is its best variant on translation.}
\label{fig:teaser}
\end{figure}
\section{Introduction}
\label{sec:intro}

Sign language translation (SLT) turns a signing video into spoken-language text.
For American Sign Language (ASL), several public corpora now exist, ranging from studio recordings such as How2Sign~\cite{duarte2021how2sign} to large web collections such as YouTube-ASL~\cite{youtubeasl2023}, OpenASL~\cite{shi2022openasl}, ASL STEM Wiki~\cite{yin2024stemwiki} and YouTube-SL-25~\cite{tanzer2025ytsl25}. Several recent methods use the larger datasets for pretraining and report gains in different test settings and representations~\cite{li2025unisign,gueuwou2025shubert}.
These approaches vary across multiple dimensions, including the underlying pose estimator or visual features, pretraining corpora, and evaluation splits, making them hard to compare. 
To address these discrepancies, we propose a novel pretraining strategy and introduce a standardized ASL corpus and evaluation protocol, allowing methods to be compared on equal footing. 

The novel pretraining approach involves masked motion modeling (MMM) on continuous keypoints in the first stage, followed by the joint masking of motion and pseudo-gloss units of the corresponding captions. 
Short gloss-like representations of the caption are obtained by a large language model (LLM). 
The pretrained encoder is then connected to a pretrained language model, T5~\cite{raffel2020t5}, and trained further for translation.

In order to enable large-scale pretraining we then create a new ASL corpus from the five public ASL datasets of about 1,100 hours, extract 2D keypoints with the RTMPose~\cite{jiang2023rtmpose} pose estimator for all clips, and record which test videos also appear in other corpora. 
Under the shared evaluation protocol, the proposed model outperforms our reproductions of Uni-Sign~\cite{li2025unisign} (with and without RGB), SHuBERT~\cite{gueuwou2025shubert} and the YouTube-ASL baseline~\cite{youtubeasl2023} on How2Sign, ASL STEM Wiki and OpenASL.
Controlled ablations show that both pretraining stages are needed, and that the benefit of joint pretraining grows as the translation data shrinks.
Our contributions are as follows.
\begin{itemize}
\item A two-stage pretraining for sign motion, masked motion modeling followed by joint motion and pseudo-gloss masked modeling. Its benefit grows as labelled data shrinks.
\item A curated ASL corpus of about 1,100 hours from five public datasets, with unified 2D keypoints, filtered erroneous captions and a record of the videos shared across corpora, so that test videos can be kept out of training or reported separately.
\item A simple translation model that connects the pretrained encoder to a pretrained language model, T5~\cite{raffel2020t5}, and is fine-tuned for translation, outperforming our reproductions of state-of-the-art baselines SHuBERT~\cite{gueuwou2025shubert} and Uni-Sign~\cite{li2025unisign} on How2Sign, ASL STEM Wiki and OpenASL. The same encoder also improves isolated sign language recognition (ISLR) on ASL Citizen and WLASL2000~\cite{li2020word}.
\end{itemize}
\noindent
By establishing a standardized training and evaluation protocol, we enable a fair comparison among existing methods, offering valuable insights into the roles of input representations, pretraining strategies, and architectures for sign language translation. 
We will release all code, the data and every model weight, so that every experiment in this paper can be rerun from scratch.

\section{Related Work}
\label{sec:related}

\paragraph{Sign language translation.}
Neural SLT started with encoder-decoder models on PHOENIX-2014T~\cite{camgoz2018nslt} focusing on joint recognition and translation with gloss supervision~\cite{camgoz2020slt}.
Glosses are expensive to annotate and absent in large datasets, so more recent approaches exploit the larger datasets and vary in the architectures, pretraining strategies and representations they use~\cite{li2025unisign,gueuwou2025shubert}. 
GFSLT-VLP~\cite{zhou2023gfslt} aligns visual features with text through CLIP-style pretraining, GloFE~\cite{lin2023glofe} uses text semantics as weak supervision, and several methods connect pretrained visual encoders, such as DINOv2~\cite{oquab2024dinov2}, to LLMs~\cite{wong2024sign2gpt,gong2024signllm}.
Pseudo-glosses derived from captions also supervise the visual encoder, as content words it learns to locate in the video~\cite{wong2024sign2gpt} or as language-model outputs used as CTC targets~\cite{guo2025pgg}.
We also derive pseudo-glosses from captions with an LLM.
Our encoder instead recovers masked pseudo-gloss units from the motion and the remaining units.

\paragraph{Large-scale ASL data.}
How2Sign~\cite{duarte2021how2sign} provides about 80 hours of instructional signing recorded in a studio.
YouTube-ASL~\cite{youtubeasl2023} and YouTube-SL-25~\cite{tanzer2025ytsl25} collect captioned signing videos from the web at a much larger scale.
OpenASL~\cite{shi2022openasl} gathers news and vlog videos, and ASL STEM Wiki~\cite{yin2024stemwiki} contains professional interpretations of Wikipedia articles.
\Cref{tab:corpus} gives the statistics of these five corpora.
WLASL~\cite{li2020word} contains isolated signs from dictionary and web videos, and ASL Citizen~\cite{desai2023aslcitizen} provides isolated signs recorded by community members.
Scaling studies show that more data, including multilingual data, improves translation~\cite{youtubeasl2023,zhang2024scaling}.
These corpora differ in recording conditions, caption style and splits, and their web videos partly overlap. Because existing methods are evaluated on different combinations of datasets and different train and test splits, direct performance comparisons remain challenging.

\paragraph{Pretraining for sign language.}
Recent translation methods pretrain on large datasets with the translation task~\cite{youtubeasl2023,li2025unisign}, in some cases after separate encoder pretraining. 

Uni-Sign~\cite{li2025unisign} feeds the output of a pose encoder into mT5~\cite{xue2021mt5} and trains the whole model end to end on translation over large-scale sign language data, optionally adding RGB hand crops.
The YouTube-ASL baseline~\cite{youtubeasl2023} feeds keypoints directly into T5~\cite{raffel2020t5} and trains it on translation.
Additionally, SSVP-SLT~\cite{rust2024ssvp} and SHuBERT~\cite{gueuwou2025shubert} pretrain the encoder before translation training.
SSVP-SLT pretrains a video encoder with masked autoencoding on anonymized web videos.
SHuBERT learns multi-stream representations of the hands, face and body by masked cluster prediction and translates with ByT5~\cite{xue2022byt5}.
Pose-based pretraining for recognition, such as SignBERT~\cite{hu2021signbert} and BEST~\cite{zhao2023best}, likewise mask hand and body poses and learn to recover them.
This is similar to masked modeling of human motion~\cite{pinyoanuntapong2024mmm}, as in MotionBERT~\cite{zhu2023motionbert}.
Our method also pretrains the encoder before translation pretraining, in two stages.
Our first stage is masked motion modeling directly on 2D keypoints.
Our second stage masks pseudo-glosses and motion in one sequence, following masked language modeling~\cite{devlin2019bert} and its joint video and text extension~\cite{sun2019videobert}, so the encoder has to relate the units to the signing before it sees any translation target.

\paragraph{Evaluation and reproducibility.}
Prior work points out that SLT results are hard to compare, because of differences in data, preprocessing and metrics, and because glosses and test sets are used inconsistently~\cite{muller2023considerations}.
Many recent SLT methods release no training code or no weights, and reported numbers often depend on unreleased pretraining data.
We therefore reproduce recent methods under a unified protocol.

\section{Method}
\label{sec:method}

Following prior insights into the impact of pretraining for SLT on large-scale corpora~\cite{youtubeasl2023,li2025unisign}, we propose a novel pretraining strategy that leverages pseudo-gloss representations derived from text captions via an LLM. 

\begin{figure*}[!t]
\centering
\resizebox{0.82\textwidth}{!}{\begin{tikzpicture}[x=1cm,y=1cm,
  wtok/.style={draw=txtc!85!black, line width=0.8pt, fill=white, minimum height=0.62cm, minimum width=1.25cm, inner sep=1pt, font=\small\sffamily, rounded corners=2pt},
  mtok/.style={draw=motc!85!black, line width=0.8pt, fill=white, minimum height=0.62cm, minimum width=0.42cm, inner sep=0pt, rounded corners=2pt},
  wmask/.style={wtok, dashed, font=\scriptsize\sffamily},
  mmask/.style={mtok, dashed, font=\scriptsize\sffamily},
  wout/.style={wtok, fill=txtc!25},
  mout/.style={mtok, fill=motc!30},
  tf/.style={draw=black!70, fill=trnc!30, rounded corners=3pt, minimum width=2.9cm, minimum height=0.9cm, font=\small\sffamily, align=center},
  rowlab/.style={font=\small\sffamily\bfseries, anchor=west},
  sublab/.style={font=\small\sffamily, anchor=west},
  note/.style={font=\scriptsize\sffamily, anchor=west, text=black!75},
  head/.style={draw=black!70, fill=black!6, rounded corners=3pt, minimum height=0.62cm, minimum width=1.5cm, inner sep=2pt, font=\scriptsize\sffamily, align=center},
]
\def\yA{0}
\node[rowlab] at (-0.1,\yA+0.98) {Stage 1, masked motion modeling};
\node[note, text=motc!80!black] at (-0.1,\yA+0.55) {motion frames};
\foreach \i/\m in {1/0,2/0,3/1,4/0,5/0,6/1,7/0,8/0,9/1,10/0,11/0,12/0} {
  \pgfmathsetmacro{\x}{0.12+0.5*(\i-1)}
  \ifnum\m=1 \node[mmask] at (\x,\yA) {?}; \else \node[mtok] at (\x,\yA) {}; \fi
}
\draw[arr] (6.0,\yA) -- (6.38,\yA);
\node[tf, text width=2.55cm, align=left] (t1) at (7.85,\yA) {Motion encoder};
\flame{9.02,\yA-0.18}
\draw[arr] (9.32,\yA) -- (9.62,\yA);
\node[head] at (10.4,\yA) {\begin{tabular}{@{}c@{}}Linear\\[-3pt]head\end{tabular}\hspace{0.4cm}};
\flame{10.82,\yA-0.18}
\draw[arr] (11.15,\yA) -- (11.4,\yA);
\node[note, text=motc!80!black] at (11.45,\yA+0.55) {recovered};
\foreach \i in {1,2,3} { \node[mout] at (11.65+0.5*\i-0.5,\yA) {}; }
\node[note] at (13.15,\yA) {L1 to the true frames};

\def\yB{-2.5}
\node[rowlab] at (-0.1,-1.12) {Stage 2, joint motion and pseudo-gloss pretraining};
\node[sublab] at (0.2,\yB+0.9) {(a) pseudo-gloss mode, motion fully visible};
\node[note, text=txtc!80!black] at (-0.1,\yB+0.55) {pseudo-gloss units};
\node[wtok] at (0.55,\yB) {police};
\node[wmask] at (1.9,\yB) {[MASK]};
\node[wtok] at (3.25,\yB) {kill};
\node[font=\small] at (4.07,\yB) {+};
\node[note, text=motc!80!black] at (4.3,\yB+0.55) {motion};
\foreach \i in {1,2,3} { \node[mtok] at (4.55+0.5*\i-0.5,\yB) {}; }
\draw[arr] (6.0,\yB) -- (6.38,\yB);
\node[tf, text width=2.55cm, align=left] at (7.85,\yB) {Motion encoder\\[-2pt]{\scriptsize from stage 1}};
\flame{9.02,\yB-0.18}
\draw[arr] (9.32,\yB) -- (9.62,\yB);
\node[head] at (10.4,\yB) {\begin{tabular}{@{}c@{}}Linear\\[-3pt]head\end{tabular}\hspace{0.4cm}};
\flame{10.82,\yB-0.18}
\draw[arr] (11.15,\yB) -- (11.4,\yB);
\node[note, text=txtc!80!black] at (11.45,\yB+0.55) {recovered};
\node[wout] at (12.1,\yB) {shoot};
\node[note] at (13.15,\yB) {cross-entropy, 15--50\% of units masked};

\def\yC{-4.55}
\node[sublab] at (0.2,\yC+0.9) {(b) motion mode, pseudo-gloss units visible};
\node[note, text=txtc!80!black] at (-0.1,\yC+0.55) {pseudo-gloss units};
\node[wtok] at (0.55,\yC) {police};
\node[wtok] at (1.9,\yC) {shoot};
\node[wtok] at (3.25,\yC) {kill};
\node[font=\small] at (4.07,\yC) {+};
\node[note, text=motc!80!black] at (4.3,\yC+0.55) {motion};
\foreach \i/\m in {1/0,2/1,3/0} {
  \pgfmathsetmacro{\x}{4.55+0.5*(\i-1)}
  \ifnum\m=1 \node[mmask] at (\x,\yC) {?}; \else \node[mtok] at (\x,\yC) {}; \fi
}
\draw[arr] (6.0,\yC) -- (6.38,\yC);
\node[tf, text width=2.55cm, align=left] at (7.85,\yC) {Motion encoder\\[-2pt]{\scriptsize from stage 1}};
\flame{9.02,\yC-0.18}
\draw[arr] (9.32,\yC) -- (9.62,\yC);
\node[head] at (10.4,\yC) {\begin{tabular}{@{}c@{}}Linear\\[-3pt]head\end{tabular}\hspace{0.4cm}};
\flame{10.82,\yC-0.18}
\draw[arr] (11.15,\yC) -- (11.4,\yC);
\node[note, text=motc!80!black] at (11.45,\yC+0.55) {recovered};
\node[mout] at (11.65,\yC) {};
\node[note] at (13.15,\yC) {L1, 30--50\% of frames masked};

\draw[-{Stealth[length=4pt]}, black!60, dashed] (15.6,\yB-0.25) to[bend left=35] node[note, right, pos=0.5, align=left] {each example\\draws one mode,\\50 / 50} (15.6,\yC+0.25);
\draw[black!25] (-0.1,-0.82) -- (17.3,-0.82);

\def\yL{-5.75}
\node[wtok, minimum width=0.42cm] at (0.12,\yL) {}; \node[note] at (0.4,\yL) {pseudo-gloss unit};
\node[mtok] at (3.3,\yL) {}; \node[note] at (3.58,\yL) {motion frame};
\node[mmask] at (5.7,\yL) {?}; \node[note] at (5.98,\yL) {masked};
\node[mout] at (7.55,\yL) {}; \node[note] at (7.83,\yL) {recovered by the model};
\flame{11.25,\yL-0.18} \node[note] at (11.55,\yL) {trained in this stage};
\end{tikzpicture}}
\caption{Two-stage pretraining of the motion encoder. Stage 1 masks motion frames and recovers them. Stage 2 feeds the pseudo-gloss units and the motion of the same clip into the same motion encoder, initialized from stage 1. Each example masks either units or frames, and the model recovers what was masked. To recover a masked unit it has to find that sign in the motion. After pretraining only the encoder is kept.}
\label{fig:pretrain}
\end{figure*}
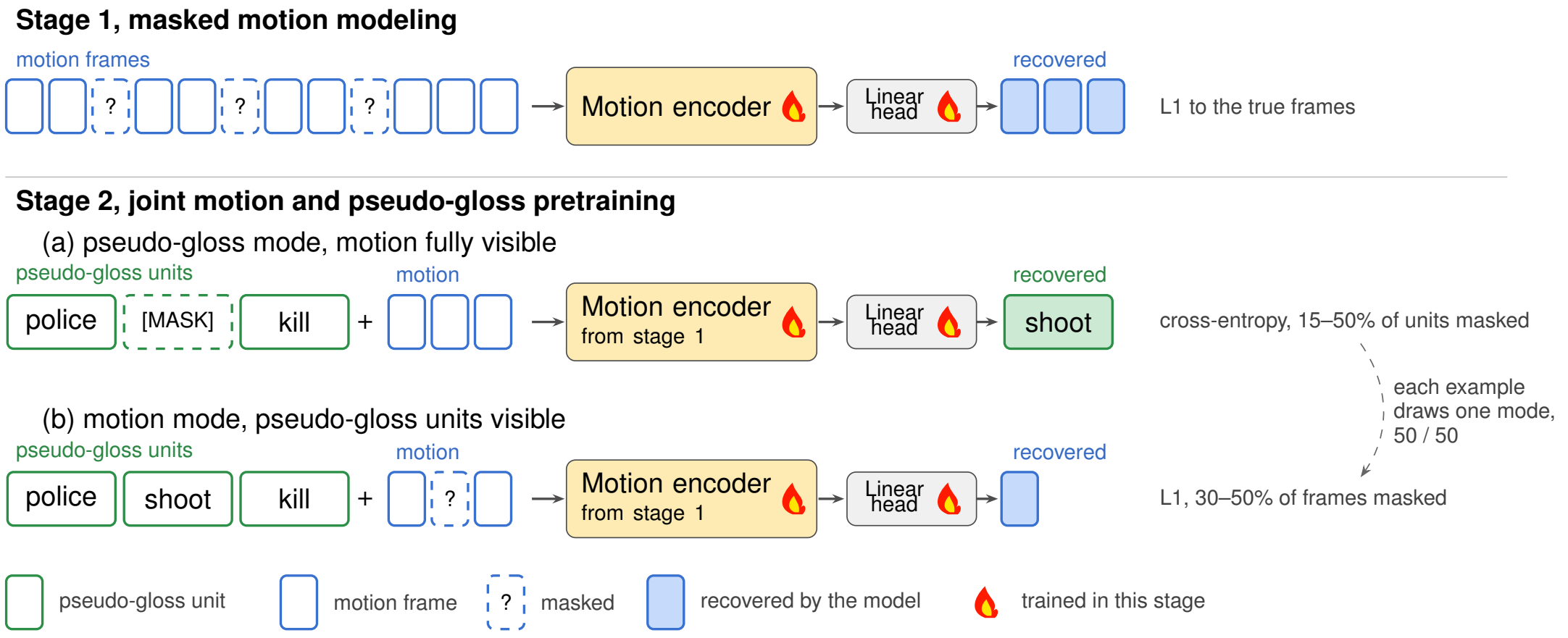

Our approach pretrains a motion encoder in two stages (\cref{fig:pretrain}) followed by connecting it to a pretrained language model for translation (\cref{fig:arch}).
The first stage of encoder pretraining, masked motion modeling (\cref{sec:mmm}), uses the motion alone. 
The second stage, joint motion and pseudo-gloss masking (\cref{sec:joint}), adds the pseudo-glosses of each caption, obtained using the LLM detailed in \cref{sec:llmgloss}.
A clip is a sequence of keypoint frames $X=(x_1,\dots,x_T)$ with $x_t\in\mathbb{R}^{207}$, which holds 69 RTMPose keypoints with $x$, $y$ and confidence.
Frames are standardized with the mean and standard deviation of the pretraining data.
The caption of a clip is a sentence $y=(y_1,\dots,y_N)$ of subword tokens.

\subsection{Motion encoder}
\label{sec:encoder}
The encoder $\mathcal{E}$ is a Transformer with $L=16$ layers, width $d=1024$ and 16 heads.
Each frame is one token.
A linear layer $W_{\mathrm{in}}$ maps $x_t$ to $\mathbb{R}^d$ and a learned position embedding $p_t$ is added, giving
\begin{equation}
h_t^{0} = W_{\mathrm{in}}\,x_t + p_t, \qquad H = \mathcal{E}(h^{0}_{1:T}) \in \mathbb{R}^{T\times d}.
\end{equation}
Clips longer than 1,024 frames are truncated.

\subsection{Masked motion modeling}
\label{sec:mmm}
We sample a set $\mathcal{M}$ with 30\% of the frame positions and replace their input embeddings by a learned vector $m\in\mathbb{R}^d$,
\begin{equation}
\tilde h_t^{0} =
\begin{cases}
m + p_t, & t\in\mathcal{M},\\
W_{\mathrm{in}}\,x_t + p_t, & t\notin\mathcal{M}.
\end{cases}
\end{equation}
With $\tilde H = \mathcal{E}(\tilde h^{0}_{1:T})$, a linear head $g$ predicts the frames, and the loss is the L1 error on the masked frames,
\begin{equation}
\mathcal{L}_{\mathrm{motion}} = \frac{1}{|\mathcal{M}|}\sum_{t\in\mathcal{M}} \big\| g(\tilde H_t) - x_t \big\|_1 .
\end{equation}
Stage 1 is trained with $\mathcal{L}_{\mathrm{motion}}$ alone.
This follows masked modeling of images and motion~\cite{he2022mae,pinyoanuntapong2024mmm}.

\subsection{Pseudo-gloss generation}
\label{sec:llmgloss}
Captions are written in English.
Their word order, function words and inflections differ from the signing, and many words are not signed at all.
For joint pretraining (\cref{sec:joint}) we therefore rewrite every caption into a short sequence of gloss-like units, which we call pseudo-glosses, with a local LLM.

\paragraph{Prompt.}
We use the open-weight LLM Qwen3.5-27B~\cite{qwen3.5} with greedy decoding, one request per unique caption (396,580 captions).
The prompt contains a sign lexicon of 3,339 entries.
It combines the glosses of ASL Signbank~\cite{hochgesang2017signbank}, a lexical database of ASL signs, with the 2,731 sign labels of ASL Citizen~\cite{desai2023aslcitizen}, and spelling variants of the same sign are merged into one entry (\cref{app:lexicon}).
The lexicon tells the model which words have an established sign, so that such a word is written as its lexicon entry and the same sign gets the same unit in every caption.
The prompt asks for one unit per sign, in the order of the caption, and asks the model to label each unit with one of the following four tags.
\begin{itemize}
\item \textbf{L}, a lexicon entry, written exactly as in the lexicon. Multi-word entries such as \emph{all day} stay one unit.
\item \textbf{W}, any other content word, written as its base form.
\item \textbf{P}, a pronoun or pointing, such as \emph{he}, \emph{this} or \emph{there}.
\item \textbf{FS}, a name or acronym that would be fingerspelled.
\end{itemize}
Articles, the copula, auxiliaries, the infinitive \emph{to} and fillers are dropped, and negation and question words are kept.
The prompt includes two worked examples.
After decoding, we check the L and W tags against the lexicon, so that a unit is tagged L exactly when it is a lexicon entry.

\paragraph{Output.}
On average a clip has 9.2 pseudo-gloss units against 14.7 caption words.
55\% of the units are lexicon entries, 29\% other content words, 10\% pronouns and 6\% fingerspelled names.
\Cref{tab:llmgloss} shows examples.
These are not real glosses and contain errors, but they are much closer to the signed content than the English caption.
For pretraining we use the pseudo-glosses without their tags.

\begin{table}[!t]
\centering
\small
\caption{Pseudo-gloss examples from the training captions. Tags are L (lexicon entry), W (other content word), P (pronoun or pointing) and FS (fingerspelled).}
\label{tab:llmgloss}
\setlength{\tabcolsep}{3pt}
\renewcommand{\arraystretch}{1.15}
\begin{tabular}{p{3.55cm}p{4.2cm}}
\toprule
Caption & Pseudo-gloss \\
\midrule
Police shot and killed the knifeman. & police$_L$ shoot$_L$ kill$_L$ knifeman$_W$ \\
People immediately started running from the school once he heard the gunfire, yelling & people$_L$ immediately$_W$ start$_L$ run$_L$ from$_L$ school$_L$ once$_L$ he$_P$ hear$_L$ gunfire$_W$ yell$_W$ \\
Hi, my name is Katie Moore. & hi$_W$ my$_P$ name$_L$ katie$_{FS}$ moore$_{FS}$ \\
He orders Moses to give them what would've belonged to their father. & he$_P$ order$_L$ moses$_{FS}$ give$_L$ they$_P$ belong$_L$ they$_P$ father$_L$ \\
The Japanese government has approved this process. & japanese$_{FS}$ government$_L$ approve$_W$ process$_L$ \\
\bottomrule
\end{tabular}
\end{table}

\subsection{Joint motion and pseudo-gloss masking}
\label{sec:joint}
We use the pseudo-gloss units of \cref{sec:llmgloss}, $u=(u_1,\dots,u_K)$ with at most 40 units per clip, and a vocabulary of the 30,000 most frequent units plus the lexicon entries.

\paragraph{Input.}
The encoder from stage 1 reads the units and the frames as one sequence.
Units are embedded with a learned table, a pseudo-gloss position embedding and a modality embedding $e_{\text{pseudo-gloss}}$, and frames as in \cref{sec:encoder} plus $e_{\mathrm{motion}}$,
\begin{equation}
Z = \mathcal{E}\big([\,\mathrm{emb}(u_{1:K}) ;\; \tilde h^{0}_{1:T}\,]\big),
\end{equation}
so every layer lets units and frames attend to each other.

\paragraph{Two masking modes.}
Each training example is drawn in one of two modes with equal probability.
In the pseudo-gloss mode we draw a fraction $r$ uniformly from $[0.15, 0.5]$, replace that fraction of the units, and at least one unit, by a [MASK] unit, and keep all frames visible.
A linear head over the unit vocabulary predicts the masked units $\mathcal{U}$,
\begin{equation}
\mathcal{L}_{\text{pseudo-gloss}} = -\frac{1}{|\mathcal{U}|}\sum_{k\in\mathcal{U}} \log P\big(u_k \mid Z_k\big).
\end{equation}
In the motion mode we draw $r$ uniformly from $[0.3, 0.5]$ and mask that fraction of the frames as in stage 1, keep the units visible (all units are removed in 10\% of the examples), and use the same loss $\mathcal{L}_{\mathrm{motion}}$ of \cref{sec:mmm} on the masked frames.
The total loss is
\begin{equation}
\mathcal{L}_{\mathrm{joint}} = \mathcal{L}_{\text{pseudo-gloss}} + \mathcal{L}_{\mathrm{motion}} .
\end{equation}

\paragraph{The encoder recovers masked pseudo-gloss units from the motion.}
The other pseudo-gloss units of a clip are usually not enough to guess a masked one, so the encoder has to find that sign in the motion.
When the motion of a validation clip is replaced by that of another clip, the accuracy on masked units drops from 0.19 to 0.01 in setting B and from 0.18 to 0.02 in setting A (\cref{app:motioncheck}), so the prediction depends on the motion.
No label tells the encoder which frames sign which unit, so it learns this mapping implicitly.
After pretraining we keep only the encoder and discard both heads.

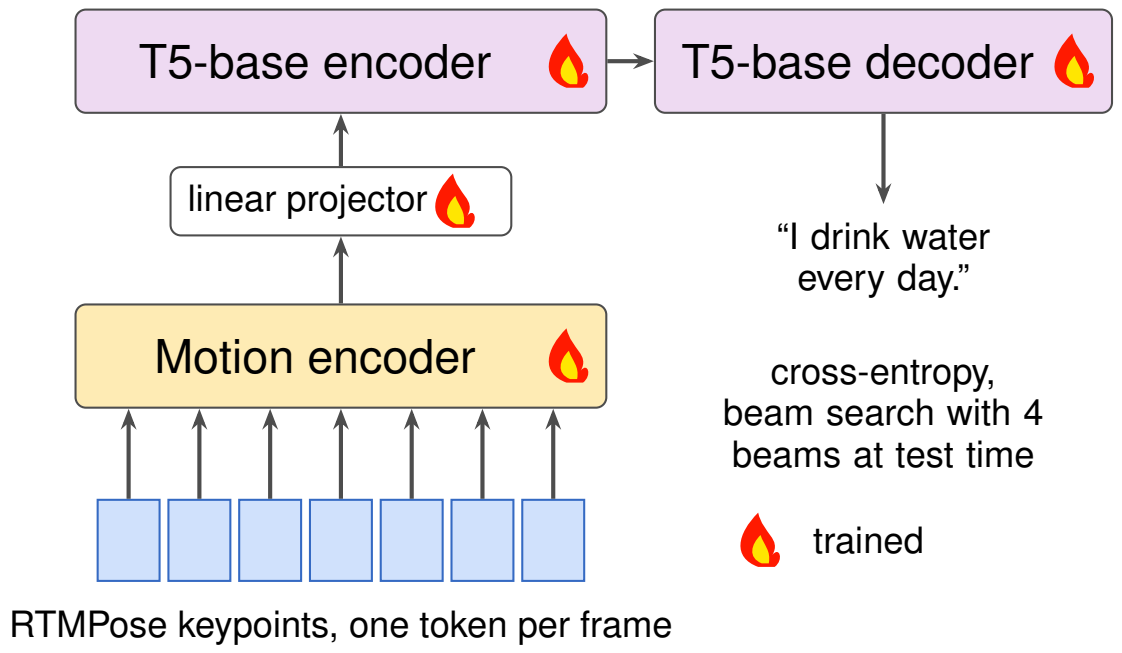
\begin{figure}[!t]
\centering
\resizebox{0.9\linewidth}{!}{\begin{tikzpicture}[x=1cm,y=1cm]
  \foreach \i in {1,...,7} { \node[frame] (f\i) at (0.48*\i,0) {}; }
  \node[lab, anchor=north] at (1.92,-0.35) {RTMPose keypoints, one token per frame};
  \node[block, minimum width=3.6cm] (enc) at (1.92,1.25) {Motion encoder\hspace{0.35cm}};
  \foreach \i in {1,...,7} { \draw[arr] (f\i.north) -- (f\i.north |- enc.south); }
  \node[block, minimum width=2.0cm, minimum height=0.45cm, fill=white] (proj) at (1.92,2.3) {\scriptsize linear projector\hspace{0.45cm}};
  \draw[arr] (enc.north) -- (proj.south);
  \node[t5block, minimum width=3.6cm] (t5e) at (1.92,3.25) {T5-base encoder\hspace{0.35cm}};
  \draw[arr] (proj.north) -- (t5e.south);
  \node[t5block, minimum width=3.1cm] (t5d) at (5.6,3.25) {T5-base decoder\hspace{0.35cm}};
  \draw[arr] (t5e.east) -- (t5d.west);
  \node[lab, text width=2.6cm] (out) at (5.6,1.9) {``I drink water\\every day.''};
  \draw[arr] (t5d.south) -- (out.north);
  \node[lab, anchor=north, text width=2.8cm] at (5.6,1.35) {cross-entropy, beam search with 4 beams at test time};
  \flame{3.45,1.07}
  \flame{2.68,2.12}
  \flame{3.45,3.07}
  \flame{6.88,3.07}
  \flame{4.75,-0.18} \node[lab, anchor=west] at (5.0,0) {trained};
\end{tikzpicture}}
\caption{Translation model. The motion encoder, initialized from both pretraining stages, is followed by a linear projector and T5-base. All parts, marked with a flame, are trained end to end.}
\label{fig:arch}
\end{figure}

\subsection{Translation}
\label{sec:translation}
For translation the pretrained encoder output $H$ is projected to the width of T5-base~\cite{raffel2020t5} and passed through the T5 encoder and decoder (\cref{fig:arch}),
\begin{equation}
P(y\mid X) = \prod_{n=1}^{N} P_{\mathrm{dec}}\big(y_n \mid y_{<n},\; \mathrm{Enc}_{\mathrm{T5}}(W_{\mathrm{p}} H)\big).
\end{equation}
All parts are trained end to end with the token cross-entropy
\begin{equation}
\mathcal{L}_{\mathrm{SLT}} = -\frac{1}{N}\sum_{n=1}^{N}\log P_{\mathrm{dec}}\big(y_n \mid y_{<n}, X\big).
\end{equation}
We keep the T5 encoder between the motion encoder and the decoder, so that the decoder reads the kind of states it was pretrained on.
At test time we decode with beam search with 4 beams.

\section{A Curated ASL Corpus}
\label{sec:corpus}

We build one corpus from five public ASL datasets, YouTube-ASL~\cite{youtubeasl2023}, YouTube-SL-25~\cite{tanzer2025ytsl25}, ASL STEM Wiki~\cite{yin2024stemwiki}, OpenASL~\cite{shi2022openasl} and How2Sign~\cite{duarte2021how2sign}.
Our goal is not a new recording effort but a single, documented processing of existing datasets, so that every method can be trained and tested on exactly the same clips.
\Cref{tab:corpus} summarizes the result.

\begin{table}[!t]
\centering
\small
\caption{The curated training corpus (setting B), after cleaning and after removing every clip that comes from the same video as a test clip. OpenASL videos partly overlap with those of YouTube-ASL and YouTube-SL-25, so the total of unique videos is lower than the sum. \Cref{tab:corpus_full} in \cref{app:corpus} gives more information, including the domain of each source and the mean duration and number of caption words per clip.}
\label{tab:corpus}
\setlength{\tabcolsep}{4pt}
\begin{tabular}{lrrr}
\toprule
Source & Videos & Clips & Hours \\
\midrule
YouTube-ASL & 8,790 & 228,399 & 440.4 \\
ASL STEM Wiki & 229 & 59,109 & 291.7 \\
OpenASL & 1,476 & 70,763 & 169.0 \\
YouTube-SL-25, ASL & 4,069 & 83,181 & 152.1 \\
How2Sign & 2,192 & 30,794 & 58.8 \\
\midrule
Total & 15,687 unique & 472,246 & 1,112.1 \\
\bottomrule
\end{tabular}
\end{table}

\subsection{Sources and clip extraction}
\paragraph{Datasets with sentence clips.}
How2Sign and ASL STEM Wiki release one video per sentence.
For How2Sign we use the official green-screen front-view clips with the re-aligned captions.
For ASL STEM Wiki (STEM) we download all 64,748 sentence recordings of the 254 articles.
OpenASL releases sentence timestamps for YouTube videos.
Some source videos are no longer available, and we cut 94,225 of the 98,420 annotated sentences.

\paragraph{Web videos with captions.}
YouTube-ASL and YouTube-SL-25 provide whole videos with caption tracks, so we cut sentence clips ourselves.
From YouTube-SL-25 we use only the ASL videos that are not already in YouTube-ASL.
We clean the captions, merge consecutive captions into sentences, and keep a sentence only when it comes from a manual caption track, has a plausible duration and signing rate, and shows a visible signer.
This keeps 245,727 YouTube-ASL clips (479 hours) and 86,551 YouTube-SL-25 clips (159 hours).
\Cref{app:clips} gives the exact rules.

\subsection{Keypoints}
\label{sec:keypoints}
All clips are processed with the same pose estimator, RTMPose~\cite{jiang2023rtmpose} in the lightweight whole-body configuration of rtmlib, which uses the RTMW model~\cite{jiang2024rtmw}, on every second video frame.
We keep the 69 body, hand and face points used by Uni-Sign~\cite{li2025unisign} with its normalization, so each frame is a 207-dimensional vector of $x$, $y$ and confidence (details in \cref{app:keypoints}).
All keypoint-based methods in this paper, ours and our reproductions, read their keypoints from this same RTMPose output, so the pose estimator is not a source of difference between methods.

\subsection{Test splits and leakage}
\label{sec:overlap}
All methods use the same fixed test lists, summarized in \cref{tab:testsets} in \cref{app:testsets}.
For How2Sign we use the train and test lists of Uni-Sign~\cite{li2025unisign} (31,044 and 2,343 clips).
ASL STEM Wiki has no official split, so we hold out 10\% of the articles (25 articles, 4,047 sentences) as test set, so that no test article is seen during training.
OpenASL uses its official split.
For setting B (\cref{sec:protocol}), where one model is trained on all five sources, we also hold out 151 YouTube-ASL videos with 3,158 sentences as a test set, and draw 2,000 validation clips from all sources, of which translation uses 600 for checkpoint selection.

The web corpora were collected independently, and several of them contain the same YouTube videos.
By YouTube identifier, 232 of the 414 OpenASL test videos are also part of YouTube-ASL.
They hold 608 of the 908 OpenASL test sentences, so any method pretrained on YouTube-ASL, which is the common practice, has seen the videos behind two thirds of the OpenASL test set.

In setting B we remove from the training set every clip that comes from the same video as a test clip.

\section{Experimental Setup}
\label{sec:setup}

\subsection{Two evaluation settings}
\label{sec:protocol}

\paragraph{YouTube-ASL pretraining and per-dataset fine-tuning (setting A).}
This follows the common practice of pretraining a translation model on YouTube-ASL and fine-tuning it on the target dataset~\cite{youtubeasl2023,li2025unisign,gueuwou2025shubert}.
Every method is first trained for translation on the same 602,243 YouTube-ASL caption segments (785 hours), for 20 epochs unless stated otherwise in \cref{app:repro}.
As in prior work~\cite{youtubeasl2023}, each segment is one YouTube-ASL caption, and we do not merge captions into sentences as in \cref{sec:corpus}.
Each method is then fine-tuned separately on How2Sign, ASL STEM Wiki and OpenASL for 20 epochs each, and we report the last checkpoint without any selection on the test set.
For our model, both pretraining stages also use only these YouTube-ASL clips.
Since YouTube-ASL contains the videos behind two thirds of the OpenASL test sentences (\cref{sec:overlap}), we report OpenASL on all 908 sentences and separately on the 300 sentences from unseen videos and the 608 sentences from seen videos.

\paragraph{One model on all data (setting B).}
One model is trained on the full curated corpus of \cref{sec:corpus} (472,246 clips, 1,112 hours).
No training clip comes from the same video as a test clip (\cref{sec:overlap}).
Both pretraining stages and the translation stage use these clips.
The model is tested on all test sets directly, without per-dataset fine-tuning.
The translation checkpoint is selected on a validation set of 600 clips drawn from all five sources.

\paragraph{Metric.}
We report BLEU-4 computed with sacreBLEU (13a tokenization, case-sensitive, default smoothing) on detokenized output.
Each number is a single run.

\subsection{Implementation of our model}
\label{sec:impl}
Stage 1 trains for 100K steps on windows of 256 frames, stage 2 for 80K steps, keeping the checkpoint with the highest masked-unit accuracy on validation clips, and translation for 20 epochs.
All stages use AdamW with a cosine schedule (\cref{tab:hparams} in \cref{app:hparams}), and the fine-tuning of setting A halves the learning rates.
All models are trained on a single A100 80\,GB GPU, in about 11 hours for stage 1, 6 to 10 hours for stage 2 and 24 to 44 hours for translation.

\subsection{Reproduced methods}
\label{sec:repro}
We reproduce three recent methods.
Uni-Sign~\cite{li2025unisign} and SHuBERT~\cite{gueuwou2025shubert} report state-of-the-art results on ASL translation, and the YouTube-ASL baseline~\cite{youtubeasl2023} is the standard reference for pretraining on YouTube-ASL.
All of them are trained on the same clips and tested on the same lists as our model, and all use beam search with 4 beams.
Every keypoint-based method, including ours, reads the same RTMPose keypoints of \cref{sec:keypoints}.
We run Uni-Sign with pose only and with its RGB hand branch.
SHuBERT releases its encoder but not its translation code, so we attach ByT5-Base~\cite{xue2022byt5} as described in the paper and report three runs, A, C1 and C2, that differ only in preprocessing and weight decay.
\Cref{app:repro} gives the full recipes and every deviation from the original papers that we are aware of.

\section{Results}
\label{sec:results}

\subsection{Comparison under the shared protocol}
\label{sec:exp_main}

\begin{table*}[!t]
\centering
\small
\caption{Translation pretraining on the same 785 hours of YouTube-ASL, then fine-tuning per dataset (setting A, BLEU-4, last checkpoint). OpenASL unseen and seen split the test set by whether the source video is in YouTube-ASL. All baselines are our reproductions under the same protocol. SHuBERT (A), (C1) and (C2) are three reproductions (\cref{app:repro}).}
\label{tab:settingA}
\begin{tabular}{llccccc}
\toprule
& & & & \multicolumn{3}{c}{OpenASL} \\
\cmidrule(lr){5-7}
Method & Input & How2Sign & STEM & all & unseen & seen \\
\midrule
YouTube-ASL baseline~\cite{youtubeasl2023} & pose & 3.32 & 3.19 & 13.23 & 6.37 & 16.10 \\
SHuBERT~\cite{gueuwou2025shubert} (A) & pose + RGB & 0.04 & 0.13 & 1.30 & 0.97 & 1.44 \\
\phantom{SHuBERT~\cite{gueuwou2025shubert}} (C1) & pose + RGB & 0.01 & 0.05 & 0.21 & 0.12 & 0.25 \\
\phantom{SHuBERT~\cite{gueuwou2025shubert}} (C2) & pose + RGB & 0.35 & 0.39 & 0.46 & 0.25 & 0.52 \\
Uni-Sign~\cite{li2025unisign} & pose & 9.28 & 19.85 & 21.56 & 15.42 & 24.19 \\
Uni-Sign~\cite{li2025unisign} & pose + RGB & 9.55 & 20.51 & 22.36 & 16.48 & 24.71 \\
\midrule
Ours, no pretraining & pose & 6.96 & 15.34 & 20.41 & 14.12 & 23.06 \\
Ours, MMM only & pose & 9.27 & 19.75 & 24.69 & 17.40 & 27.77 \\
Ours, joint only & pose & 8.43 & 17.95 & 24.28 & 15.19 & 27.99 \\
Ours, MMM + joint & pose & \textbf{9.79} & \textbf{22.00} & \textbf{27.84} & \textbf{18.92} & \textbf{31.66} \\
\bottomrule
\end{tabular}
\end{table*}

\Cref{tab:settingA} compares all methods in setting A.
Our full model is best on all three datasets.
It is above Uni-Sign with RGB by 0.2 BLEU-4 on How2Sign, 1.5 on STEM and 5.5 on OpenASL, although it reads keypoints only.
On OpenASL the gap holds on both the unseen and the seen videos (18.92 against 16.48, and 31.66 against 24.71), so it is not explained by the contamination.
Adding RGB to Uni-Sign helps by 0.3 to 0.8 BLEU-4.
The YouTube-ASL baseline learns OpenASL but stays near 3 BLEU-4 on How2Sign and STEM.

None of the three SHuBERT reproductions reaches 1.5 BLEU-4.
C1 produces the single sentence ``Thank you for watching.'' for all 2,343 How2Sign test clips, and C2 produces 7 distinct sentences on How2Sign.
Under the same 20-epoch budget, Uni-Sign and our model reach 14.4 and 18.4 validation BLEU-4, so the budget is sufficient for the other methods, and more training epochs do not help SHuBERT either (\cref{app:convergence}).

\begin{table}[!t]
\centering
\small
\setlength{\tabcolsep}{3.5pt}
\caption{One model trained on all 1,112 hours and tested without fine-tuning (setting B, BLEU-4). Data is the share of translation sentences. For 50\% and 25\% only the translation stage is trained on the subset, and the pretrained encoders are the same. All methods, including Uni-Sign, read the same pose input from RTMPose. YT-ASL denotes our held-out YouTube-ASL split.}
\label{tab:settingB}
\resizebox{\linewidth}{!}{%
\begin{tabular}{llcccc}
\toprule
Data & Method & STEM & OpenASL & How2Sign & YT-ASL \\
\midrule
\multirow{5}{*}{100\%} & Uni-Sign~\cite{li2025unisign} & 3.43 & 9.57 & 3.35 & 6.33 \\
& Ours, no pretraining & 0.76 & 4.76 & 1.77 & 2.88 \\
& Ours, joint only & 18.74 & 20.29 & 8.53 & 13.57 \\
& Ours, MMM only & 21.20 & 21.11 & 9.98 & 15.19 \\
& Ours, MMM + joint & \textbf{21.99} & \textbf{21.64} & \textbf{10.38} & \textbf{15.81} \\
\midrule
\multirow{2}{*}{50\%} & Ours, MMM only & 9.59 & 13.22 & 5.12 & 9.74 \\
& Ours, MMM + joint & \textbf{16.80} & \textbf{19.21} & \textbf{8.37} & \textbf{13.06} \\
\midrule
\multirow{2}{*}{25\%} & Ours, MMM only & 0.58 & 3.95 & 1.49 & 2.07 \\
& Ours, MMM + joint & \textbf{10.55} & \textbf{13.75} & \textbf{5.78} & \textbf{10.04} \\
\bottomrule
\end{tabular}}
\end{table}

\Cref{tab:settingB} shows setting B.
A single model trained on all data attains 22.0 BLEU-4 on STEM and 21.6 on OpenASL without any fine-tuning.
Uni-Sign, trained on the same data with the recipe of its pretraining stage in setting A and tested without fine-tuning, attains 3.4 on STEM, against 19.9 when it is fine-tuned per dataset in setting A.
Without encoder pretraining our network fails on every test set, with 0.76 BLEU-4 on STEM.
Earlier work reports similar failures on How2Sign without pretraining~\cite{zhang2026lslm,hong2026collapse}.

\subsection{Pretraining ablation}
\label{sec:exp_ablation}
The four rows of our model in \cref{tab:settingA} and in the 100\% block of \cref{tab:settingB} differ only in how the encoder is pretrained, and the translation training after it is identical.
Stage 1 is masked motion modeling (MMM), and stage 2 is joint motion and pseudo-gloss pretraining (joint).
No pretraining starts the encoder from random weights, MMM only uses the encoder after stage 1, joint only runs stage 2 from random weights for 180K steps, as many as both stages together, with the stage-1 learning rate, and MMM + joint is our full model.
MMM gives most of the gain over no pretraining, for example 15.34 to 19.75 BLEU-4 on STEM in setting A and 0.76 to 21.20 in setting B.
Joint pretraining on top of it adds 0.5 to 3.2 BLEU-4 in setting A and 0.4 to 0.8 BLEU-4 in setting B.
Joint pretraining alone is worse than MMM alone on almost every test set in both settings, although it trains for as many steps, so its gain does not come from longer training.
It needs a motion representation to start from, and the two stages are complementary.

\subsection{What joint pretraining changes in the encoder}
\label{sec:exp_tsne}
\begin{figure}[!t]
\centering
\includegraphics[width=\linewidth]{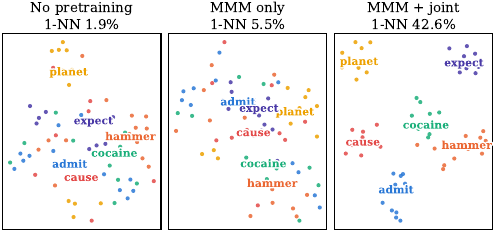}
\caption{t-SNE of the encoder features of ASL Citizen test clips from unseen signers, for six randomly chosen signs, with the setting-A encoders before any translation or recognition training. Each point is one clip. The title gives the share of all 32,928 test clips whose nearest clip from another signer is the same sign (chance 0.04\%).}
\label{fig:tsne}
\end{figure}

To see what the encoder learns in each pretraining stage, we take the three setting-A encoders, which are trained on YouTube-ASL only, and encode the 32,928 test clips of ASL Citizen, recorded by 11 signers that none of these encoders has seen.
We use no fine-tuning, and each clip is represented by the average encoder feature over its frames.
For each clip, we look among the clips of the other signers and take the one with the closest feature.
If this closest clip shows the same sign, we count it as correct.
The share of correct clips is 1.9\% without pretraining, 5.5\% after MMM, and 42.6\% after MMM and joint pretraining.
Random guessing over the 2,731 signs gives 0.04\%.
\Cref{fig:tsne} shows the clips of six randomly chosen signs.
After joint pretraining, clips of the same sign form clear clusters across signers.
The encoders never see ASL Citizen videos during pretraining, and the sign names only enter stage 2 as pseudo-gloss units of YouTube-ASL captions.
Joint pretraining therefore ties these units to their signs in continuous signing well enough that isolated signs by new signers fall together, which is what translation and recognition later build on.

\subsection{Translation data size}
\label{sec:exp_datasize}
In setting B the translation data is large, which may make the effect of pretraining less visible.
We therefore train the translation stage on random 50\% and 25\% subsets of the training sentences, drawn once with a fixed seed so that the 25\% subset is part of the 50\% subset.
The training length is kept at 20 epochs, so the number of steps shrinks with the data, and the encoders are the same pretrained ones as before.
The lower blocks of \cref{tab:settingB} show that the gap between the two encoders grows as the data shrinks, from 0.4 to 0.8 BLEU-4 with all data to 3.3 to 7.2 with half and 4.3 to 10.0 with a quarter.
These results separate the roles of the two stages.
With all data, masked motion modeling alone already gives most of the gain over no pretraining (\cref{tab:settingB}).
Its benefit drops quickly as the translation data shrinks, to 9.59 and 0.58 BLEU-4 on STEM with half and a quarter of the data, so the MMM-only encoder depends on the large amount of paired data that our curated corpus provides.
Joint pretraining keeps the encoder useful when paired data is scarce, and with a quarter of the data it still reaches 10.55 BLEU-4 on STEM, where masked motion modeling alone fails.
The validation curves in \cref{app:datasize} show that the jointly pretrained encoder is ahead from the first evaluation.

\subsection{Isolated sign recognition}
\label{sec:exp_aslc}
To test the encoder outside translation, we train a classifier on ASL Citizen~\cite{desai2023aslcitizen}.
We use the official signer-disjoint split released with the dataset, with 2,731 signs, 40,149 training clips, 10,304 validation clips and 32,928 test clips (13 test clips shorter than two keypoint frames are dropped).
The clips are processed with the same RTMPose keypoints as our corpus, and each clip is trimmed to the span in which a hand is detected.
The encoder is followed by masked mean pooling, a layer normalization, dropout and a linear layer.
We train for 40 epochs with batch 128, AdamW, learning rate 3e-4 for all parameters, label smoothing 0.1 and random temporal crops, select the epoch on the validation set and report the test accuracy.
The rows of \cref{tab:aslc} differ only in how the encoder is pretrained.
We use the three setting-B encoders.
Pretraining raises top-1 accuracy from 61.1\% to 74.3\% with MMM, and joint pretraining adds another 3.7 points to 78.0\%.
The gain of joint pretraining is larger here than in translation with all data, which matches the data-size study, since ASL Citizen has only about 15 training clips per sign.
Our classifier with the jointly pretrained encoder is the best on the official split, 2.9 points above PoseFormer~\cite{vandendriessche2025oneshot} (75.1\%), the best published result we found, whose split is not stated.
It is also above I3D, SHuBERT and SignDEG, which read RGB.

\begin{table}[!t]
\centering
\small
\setlength{\tabcolsep}{3pt}
\caption{ASL Citizen isolated sign recognition, 2,731 classes, official signer-disjoint split released with the dataset (accuracy in \%). Published results are taken from the cited papers, where recall@1 equals top-1 accuracy. $^\dagger$The paper does not state which split it evaluates on. An ensemble method~\cite{alkhoraif2025ensemble} reports 86.5\% top-1 on a subset of the 100 most frequent signs of ASL Citizen, not on all 2,731 signs, so it cannot be compared directly. Our three rows use the same classifier and recipe and differ only in how the encoder is pretrained.}
\label{tab:aslc}
\begin{tabular}{llcc}
\toprule
Method & Input & Top-1 & Top-5 \\
\midrule
ST-GCN~\cite{desai2023aslcitizen} & pose & 60 & 83 \\
SignCLIP~\cite{jiang2024signclip} & pose & 60 & 84 \\
PoseFormer$^\dagger$~\cite{vandendriessche2025oneshot} & pose & 75.1 & 93.2 \\
I3D~\cite{desai2023aslcitizen} & RGB & 63 & 86 \\
SHuBERT~\cite{gueuwou2025shubert} & pose + RGB & 65 & 87 \\
SignDEG~\cite{hu2026signdeg} & RGB & 70.4 & 89 \\
\midrule
Ours, no pretraining & pose & 61.1 & 83.6 \\
Ours, MMM only & pose & 74.3 & 91.6 \\
Ours, MMM + joint & pose & \textbf{78.0} & \textbf{94.0} \\
\bottomrule
\end{tabular}
\end{table}

We also evaluate on WLASL~\cite{li2020word} with its official split, on the 100-sign and 2,000-sign subsets (WLASL2000 has 14,289 training, 3,916 validation and 2,878 test clips).
We use the official release of the videos, in which all annotated clips are available, and extract the same RTMPose keypoints.
One training clip whose annotated span lies outside its video is dropped for all methods.
Our classifier uses the MMM + joint encoder of setting B and the same recipe as on ASL Citizen.
Uni-Sign reports state-of-the-art results on WLASL~\cite{li2025unisign}, so we compare against it only, and we reproduce it under the same conditions.
We fine-tune our setting-B Uni-Sign model, which was trained on the same 1,112 hours, with the released WLASL recipe of Uni-Sign (20 epochs, batch 32, learning rate 3e-4, at most 64 frames, the sign name generated as text and scored by exact match).
Both methods select the epoch on the validation set and report per-instance top-1 accuracy on the test set.
\Cref{tab:wlasl} shows that our encoder is 4.9 points above Uni-Sign on WLASL2000, while Uni-Sign is 0.8 points ahead on WLASL100, whose test set has only 258 clips.

\begin{table}[!t]
\centering
\small
\caption{WLASL, official split, per-instance top-1 accuracy (\%). Both methods start from training on the same 1,112 hours and read the same RTMPose keypoints. Uni-Sign is our reproduction with its released WLASL recipe.}
\label{tab:wlasl}
\begin{tabular}{llcc}
\toprule
Method & Input & WLASL100 & WLASL2000 \\
\midrule
Uni-Sign~\cite{li2025unisign} & pose & \textbf{84.5} & 47.3 \\
Ours, MMM + joint & pose & 83.7 & \textbf{52.2} \\
\bottomrule
\end{tabular}
\end{table}

\section{Conclusion}
\label{sec:conclusion}
We proposed a novel motion encoder trained on 2D pose keypoints with a two-stage pretraining strategy, masked motion modeling followed by joint motion and pseudo-gloss masking.
The encoder is then connected to a pretrained language model and fine-tuned for translation on multiple datasets.
To enable large-scale pretraining and a fair comparison with state-of-the-art methods, we curated and preprocessed about 1,100 hours of public ASL data into one corpus with shared keypoints.
By reproducing state-of-the-art translation methods on three translation benchmarks, we demonstrated the efficacy of our pretraining approach, which achieves the best performance on How2Sign, ASL STEM Wiki and OpenASL.
The encoder alone also improves isolated sign recognition on ASL Citizen and WLASL2000.
An ablation study shows that the joint pretraining stage is most effective when translation data is limited.
We will release the corpus, the code and the weights of the models we reproduced in this paper.

\paragraph{Limitations.}
Our models read 2D keypoints only, and combining them with RGB or 3D input is left to future work.
The pseudo-gloss units are produced by an LLM rather than annotated by linguists, and better units may further improve joint pretraining.

\FloatBarrier
{\small
\bibliographystyle{ieeenat_fullname}
\bibliography{references}

\begin{thebibliography}{40}
\providecommand{\natexlab}[1]{#1}
\providecommand{\url}[1]{\texttt{#1}}
\expandafter\ifx\csname urlstyle\endcsname\relax
  \providecommand{\doi}[1]{doi: #1}\else
  \providecommand{\doi}{doi: \begingroup \urlstyle{rm}\Url}\fi

\bibitem[Alkhoraif et~al.(2026)Alkhoraif, Alsulaiman, Abdul, and
  Bencherif]{alkhoraif2025ensemble}
Abdulelah~Ali Alkhoraif, Mansour Alsulaiman, Wadood Abdul, and Mohamed
  Bencherif.
\newblock Ensemble transformer-based word-level sign language recognition with
  multi-modal input fusion.
\newblock \emph{Journal of Engineering Research}, 14\penalty0 (1):\penalty0
  738--747, 2026.

\bibitem[Camgoz et~al.(2018)Camgoz, Hadfield, Koller, Ney, and
  Bowden]{camgoz2018nslt}
Necati~Cihan Camgoz, Simon Hadfield, Oscar Koller, Hermann Ney, and Richard
  Bowden.
\newblock Neural sign language translation.
\newblock In \emph{IEEE/CVF Conference on Computer Vision and Pattern
  Recognition (CVPR)}, 2018.

\bibitem[Camgoz et~al.(2020)Camgoz, Koller, Hadfield, and
  Bowden]{camgoz2020slt}
Necati~Cihan Camgoz, Oscar Koller, Simon Hadfield, and Richard Bowden.
\newblock Sign language transformers: Joint end-to-end sign language
  recognition and translation.
\newblock In \emph{Proceedings of the IEEE/CVF Conference on Computer Vision
  and Pattern Recognition (CVPR)}, 2020.

\bibitem[Desai et~al.(2023)Desai, Berger, Minakov, Milan, Singh, Pumphrey,
  Ladner, Daum{\'e}~III, Lu, Caselli, and Bragg]{desai2023aslcitizen}
Aashaka Desai, Lauren Berger, Fyodor~O. Minakov, Vanessa Milan, Chinmay Singh,
  Kriston Pumphrey, Richard~E. Ladner, Hal Daum{\'e}~III, Alex~X. Lu, Naomi
  Caselli, and Danielle Bragg.
\newblock {ASL Citizen}: A community-sourced dataset for advancing isolated
  sign language recognition.
\newblock In \emph{Advances in Neural Information Processing Systems (NeurIPS)
  Datasets and Benchmarks Track}, 2023.

\bibitem[Devlin et~al.(2019)Devlin, Chang, Lee, and Toutanova]{devlin2019bert}
Jacob Devlin, Ming-Wei Chang, Kenton Lee, and Kristina Toutanova.
\newblock {BERT}: Pre-training of deep bidirectional transformers for language
  understanding.
\newblock In \emph{Conference of the North American Chapter of the Association
  for Computational Linguistics (NAACL)}, 2019.

\bibitem[Duarte et~al.(2021)Duarte, Palaskar, Ventura, Ghadiyaram, DeHaan,
  Metze, Torres, and Giro-i Nieto]{duarte2021how2sign}
Amanda Duarte, Shruti Palaskar, Lucas Ventura, Deepti Ghadiyaram, Kenneth
  DeHaan, Florian Metze, Jordi Torres, and Xavier Giro-i Nieto.
\newblock How2sign: A large-scale multimodal dataset for continuous american
  sign language.
\newblock In \emph{Proceedings of the IEEE/CVF Conference on Computer Vision
  and Pattern Recognition (CVPR)}, 2021.

\bibitem[Gong et~al.(2024)Gong, Foo, He, Rahmani, and Liu]{gong2024signllm}
Jia Gong, Lin~Geng Foo, Yixuan He, Hossein Rahmani, and Jun Liu.
\newblock {LLMs} are good sign language translators.
\newblock In \emph{IEEE/CVF Conference on Computer Vision and Pattern
  Recognition (CVPR)}, 2024.

\bibitem[Gueuwou et~al.(2025)Gueuwou, Du, Shakhnarovich, Livescu, and
  Liu]{gueuwou2025shubert}
Shester Gueuwou, Xiaodan Du, Greg Shakhnarovich, Karen Livescu, and
  Alexander~H. Liu.
\newblock {SHuBERT}: Self-supervised sign language representation learning via
  multi-stream cluster prediction.
\newblock In \emph{Annual Meeting of the Association for Computational
  Linguistics (ACL)}, 2025.

\bibitem[Guo et~al.(2025)Guo, Li, and Cohn]{guo2025pgg}
Jianyuan Guo, Peike Li, and Trevor Cohn.
\newblock Bridging sign and spoken languages: Pseudo gloss generation for sign
  language translation.
\newblock \emph{Advances in Neural Information Processing Systems},
  38:\penalty0 77471--77499, 2025.

\bibitem[He et~al.(2022)He, Chen, Xie, Li, Doll{\'a}r, and Girshick]{he2022mae}
Kaiming He, Xinlei Chen, Saining Xie, Yanghao Li, Piotr Doll{\'a}r, and Ross
  Girshick.
\newblock Masked autoencoders are scalable vision learners.
\newblock In \emph{IEEE/CVF Conference on Computer Vision and Pattern
  Recognition (CVPR)}, 2022.

\bibitem[Hochgesang et~al.(2017)Hochgesang, Crasborn, and
  Lillo-Martin]{hochgesang2017signbank}
Julie~A. Hochgesang, Onno Crasborn, and Diane Lillo-Martin.
\newblock {ASL Signbank}.
\newblock \url{https://aslsignbank.com}, 2017.

\bibitem[Hong and Ko{\v{s}}eck{\'a}(2026)]{hong2026collapse}
Rui Hong and Jana Ko{\v{s}}eck{\'a}.
\newblock Conditional collapse in sign language production: A diagnostic and a
  coverage argument.
\newblock arXiv:2606.01643, 2026.

\bibitem[Hu et~al.(2021)Hu, Zhao, Zhou, Wang, and Li]{hu2021signbert}
Hezhen Hu, Weichao Zhao, Wengang Zhou, Yuechen Wang, and Houqiang Li.
\newblock {SignBERT}: Pre-training of hand-model-aware representation for sign
  language recognition.
\newblock In \emph{IEEE/CVF International Conference on Computer Vision
  (ICCV)}, 2021.

\bibitem[Hu et~al.(2026)Hu, He, Huang, Li, Yuan, Li, and Fang]{hu2026signdeg}
Junyi Hu, Zhewen He, Haomian Huang, Zhenhua Li, Shuaihang Yuan, Zhifei Li, and
  Yi Fang.
\newblock {DINO} eats geometry: Self-supervised sign representations that
  absorb articulated hand geometry.
\newblock \emph{arXiv preprint arXiv:2609.06296}, 2026.

\bibitem[Jiang et~al.(2023)Jiang, Lu, Zhang, Ma, Han, Lyu, Li, and
  Chen]{jiang2023rtmpose}
Tao Jiang, Peng Lu, Li Zhang, Ningsheng Ma, Rui Han, Chengqi Lyu, Yining Li,
  and Kai Chen.
\newblock {RTMPose}: Real-time multi-person pose estimation based on {MMPose}.
\newblock \emph{arXiv preprint arXiv:2303.07399}, 2023.

\bibitem[Jiang et~al.(2024{\natexlab{a}})Jiang, Xie, and Li]{jiang2024rtmw}
Tao Jiang, Xinchen Xie, and Yining Li.
\newblock {RTMW}: Real-time multi-person {2D} and {3D} whole-body pose
  estimation.
\newblock \emph{arXiv preprint arXiv:2407.08634}, 2024{\natexlab{a}}.

\bibitem[Jiang et~al.(2024{\natexlab{b}})Jiang, Sant, Moryossef, M{\"u}ller,
  Sennrich, and Ebling]{jiang2024signclip}
Zifan Jiang, Gerard Sant, Amit Moryossef, Mathias M{\"u}ller, Rico Sennrich,
  and Sarah Ebling.
\newblock {SignCLIP}: Connecting text and sign language by contrastive
  learning.
\newblock In \emph{Conference on Empirical Methods in Natural Language
  Processing (EMNLP)}, 2024{\natexlab{b}}.

\bibitem[Li et~al.(2020)Li, Rodriguez, Yu, and Li]{li2020word}
Dongxu Li, Cristian Rodriguez, Xin Yu, and Hongdong Li.
\newblock Word-level deep sign language recognition from video: A new
  large-scale dataset and methods comparison.
\newblock In \emph{The IEEE Winter Conference on Applications of Computer
  Vision}, pages 1459--1469, 2020.

\bibitem[Li et~al.(2025)Li, Zhou, Zhao, Wu, Hu, and Li]{li2025unisign}
Zecheng Li, Wengang Zhou, Weichao Zhao, Kepeng Wu, Hezhen Hu, and Houqiang Li.
\newblock Uni-sign: Toward unified sign language understanding at scale.
\newblock In \emph{International Conference on Learning Representations
  (ICLR)}, 2025.

\bibitem[Lin et~al.(2023)Lin, Wang, Zhu, Sun, Zhang, and Yang]{lin2023glofe}
Kezhou Lin, Xiaohan Wang, Linchao Zhu, Ke Sun, Bang Zhang, and Yi Yang.
\newblock Gloss-free end-to-end sign language translation.
\newblock In \emph{Annual Meeting of the Association for Computational
  Linguistics (ACL)}, 2023.

\bibitem[M{\"u}ller et~al.(2023)M{\"u}ller, Jiang, Moryossef, Rios, and
  Ebling]{muller2023considerations}
Mathias M{\"u}ller, Zifan Jiang, Amit Moryossef, Annette Rios, and Sarah
  Ebling.
\newblock Considerations for meaningful sign language machine translation based
  on glosses.
\newblock In \emph{Annual Meeting of the Association for Computational
  Linguistics (ACL)}, 2023.

\bibitem[Oquab et~al.(2024)Oquab, Darcet, Moutakanni, Vo, Szafraniec, Khalidov,
  Fernandez, Haziza, Massa, El-Nouby, et~al.]{oquab2024dinov2}
Maxime Oquab, Timoth{\'e}e Darcet, Th{\'e}o Moutakanni, Huy Vo, Marc
  Szafraniec, Vasil Khalidov, Pierre Fernandez, Daniel Haziza, Francisco Massa,
  Alaaeldin El-Nouby, et~al.
\newblock {DINOv2}: Learning robust visual features without supervision.
\newblock \emph{Transactions on Machine Learning Research}, 2024.

\bibitem[Pinyoanuntapong et~al.(2024)Pinyoanuntapong, Wang, Lee, and
  Chen]{pinyoanuntapong2024mmm}
Ekkasit Pinyoanuntapong, Pu Wang, Minwoo Lee, and Chen Chen.
\newblock {MMM}: Generative masked motion model.
\newblock In \emph{IEEE/CVF Conference on Computer Vision and Pattern
  Recognition (CVPR)}, 2024.

\bibitem[{Qwen Team}(2026)]{qwen3.5}
{Qwen Team}.
\newblock {Qwen3.5}: Towards native multimodal agents, 2026.

\bibitem[Raffel et~al.(2020)Raffel, Shazeer, Roberts, Lee, Narang, Matena,
  Zhou, Li, and Liu]{raffel2020t5}
Colin Raffel, Noam Shazeer, Adam Roberts, Katherine Lee, Sharan Narang, Michael
  Matena, Yanqi Zhou, Wei Li, and Peter~J. Liu.
\newblock Exploring the limits of transfer learning with a unified text-to-text
  transformer.
\newblock \emph{Journal of Machine Learning Research}, 21\penalty0
  (140):\penalty0 1--67, 2020.

\bibitem[Rust et~al.(2024)Rust, Shi, Wang, Camgoz, and Maillard]{rust2024ssvp}
Phillip Rust, Bowen Shi, Skyler Wang, Necati~Cihan Camgoz, and Jean Maillard.
\newblock Towards privacy-aware sign language translation at scale.
\newblock In \emph{Annual Meeting of the Association for Computational
  Linguistics (ACL)}, 2024.

\bibitem[Shi et~al.(2022)Shi, Brentari, Shakhnarovich, and
  Livescu]{shi2022openasl}
Bowen Shi, Diane Brentari, Greg Shakhnarovich, and Karen Livescu.
\newblock Open-domain sign language translation learned from online video.
\newblock In \emph{Conference on Empirical Methods in Natural Language
  Processing (EMNLP)}, 2022.

\bibitem[Sun et~al.(2019)Sun, Myers, Vondrick, Murphy, and
  Schmid]{sun2019videobert}
Chen Sun, Austin Myers, Carl Vondrick, Kevin Murphy, and Cordelia Schmid.
\newblock {VideoBERT}: A joint model for video and language representation
  learning.
\newblock In \emph{IEEE/CVF International Conference on Computer Vision
  (ICCV)}, 2019.

\bibitem[Tanzer and Zhang(2025)]{tanzer2025ytsl25}
Garrett Tanzer and Biao Zhang.
\newblock {YouTube-SL-25}: A large-scale, open-domain multilingual sign
  language parallel corpus.
\newblock In \emph{International Conference on Learning Representations
  (ICLR)}, 2025.

\bibitem[Uthus et~al.(2023)Uthus, Tanzer, and Georg]{youtubeasl2023}
Dave Uthus, Garrett Tanzer, and Manfred Georg.
\newblock Youtube-asl: A large-scale, open-domain american sign
  language-english parallel corpus.
\newblock In \emph{Advances in Neural Information Processing Systems
  (NeurIPS)}, 2023.

\bibitem[Vandendriessche et~al.(2025)Vandendriessche, De~Coster, Lejon, and
  Dambre]{vandendriessche2025oneshot}
Toon Vandendriessche, Mathieu De~Coster, Annelies Lejon, and Joni Dambre.
\newblock Representing signs as signs: One-shot {ISLR} to facilitate functional
  sign language technologies.
\newblock \emph{arXiv preprint arXiv:2502.20171}, 2025.

\bibitem[Wong et~al.(2024)Wong, Camgoz, and Bowden]{wong2024sign2gpt}
Ryan Wong, Necati~Cihan Camgoz, and Richard Bowden.
\newblock {Sign2GPT}: Leveraging large language models for gloss-free sign
  language translation.
\newblock In \emph{International Conference on Learning Representations
  (ICLR)}, 2024.

\bibitem[Xue et~al.(2021)Xue, Constant, Roberts, Kale, Al-Rfou, Siddhant,
  Barua, and Raffel]{xue2021mt5}
Linting Xue, Noah Constant, Adam Roberts, Mihir Kale, Rami Al-Rfou, Aditya
  Siddhant, Aditya Barua, and Colin Raffel.
\newblock {mT5}: A massively multilingual pre-trained text-to-text transformer.
\newblock In \emph{Conference of the North American Chapter of the Association
  for Computational Linguistics (NAACL)}, 2021.

\bibitem[Xue et~al.(2022)Xue, Barua, Constant, Al-Rfou, Narang, Kale, Roberts,
  and Raffel]{xue2022byt5}
Linting Xue, Aditya Barua, Noah Constant, Rami Al-Rfou, Sharan Narang, Mihir
  Kale, Adam Roberts, and Colin Raffel.
\newblock {ByT5}: Towards a token-free future with pre-trained byte-to-byte
  models.
\newblock \emph{Transactions of the Association for Computational Linguistics},
  2022.

\bibitem[Yin et~al.(2024)Yin, Singh, Minakov, Milan, Daum{\'e}~III, Zhang, Lu,
  and Bragg]{yin2024stemwiki}
Kayo Yin, Chinmay Singh, Fyodor~O. Minakov, Vanessa Milan, Hal Daum{\'e}~III,
  Cyril Zhang, Alex~X. Lu, and Danielle Bragg.
\newblock {ASL STEM Wiki}: Dataset and benchmark for interpreting {STEM}
  articles.
\newblock In \emph{Conference on Empirical Methods in Natural Language
  Processing (EMNLP)}, 2024.

\bibitem[Zhang et~al.(2024)Zhang, Tanzer, and Firat]{zhang2024scaling}
Biao Zhang, Garrett Tanzer, and Orhan Firat.
\newblock Scaling sign language translation.
\newblock In \emph{Advances in Neural Information Processing Systems
  (NeurIPS)}, 2024.

\bibitem[Zhang et~al.(2026)Zhang, He, Liu, Xia, Zhao, Tan, Li, Liu, Metaxas,
  and Kapadia]{zhang2026lslm}
Sen Zhang, Xiaoxiao He, Di Liu, Zhaoyang Xia, Mingyu Zhao, Chaowei Tan, Vivian
  Li, Bo Liu, Dimitris~N. Metaxas, and Mubbasir Kapadia.
\newblock Large sign language models: Toward {3D} {American Sign Language}
  translation.
\newblock In \emph{Proceedings of the IEEE/CVF Winter Conference on
  Applications of Computer Vision (WACV)}, 2026.

\bibitem[Zhao et~al.(2023)Zhao, Hu, Zhou, Shi, and Li]{zhao2023best}
Weichao Zhao, Hezhen Hu, Wengang Zhou, Jiaxin Shi, and Houqiang Li.
\newblock {BEST}: {BERT} pre-training for sign language recognition with
  coupling tokenization.
\newblock In \emph{AAAI Conference on Artificial Intelligence}, 2023.

\bibitem[Zhou et~al.(2023)Zhou, Chen, Clap{\'e}s, Wan, Liang, Escalera, Lei,
  and Zhang]{zhou2023gfslt}
Benjia Zhou, Zhigang Chen, Albert Clap{\'e}s, Jun Wan, Yanyan Liang, Sergio
  Escalera, Zhen Lei, and Du Zhang.
\newblock Gloss-free sign language translation: Improving from visual-language
  pretraining.
\newblock In \emph{IEEE/CVF International Conference on Computer Vision
  (ICCV)}, 2023.

\bibitem[Zhu et~al.(2023)Zhu, Ma, Liu, Liu, Wu, and Wang]{zhu2023motionbert}
Wentao Zhu, Xiaoxuan Ma, Zhaoyang Liu, Libin Liu, Wayne Wu, and Yizhou Wang.
\newblock {MotionBERT}: A unified perspective on learning human motion
  representations.
\newblock In \emph{IEEE/CVF International Conference on Computer Vision
  (ICCV)}, 2023.

\end{thebibliography}
}
\clearpage
\setcounter{page}{1}
\maketitlesupplementary
\appendix
\section{Caption cleaning and clip selection}
\label{app:clips}
This section gives the rules used to cut sentence clips from YouTube-ASL and YouTube-SL-25 (\cref{sec:corpus}).
We downloaded 10,709 YouTube-ASL videos, and 5,325 of the 6,015 ASL videos of YouTube-SL-25 that are not already in YouTube-ASL.
For YouTube-SL-25 we take the English caption track, falling back to tracks labelled as ASL when they contain English text.
Captions are first cleaned, by removing markup, bracketed tags, speaker labels and credit lines.
Consecutive captions are then merged into sentences until one ends with a sentence mark, the gap to the next caption exceeds 1 second, or the span exceeds 15 seconds or 30 words.
We keep a sentence as a clip when it lasts between 0.5 and 30 seconds, comes from a manual caption track rather than automatic speech recognition, starts with a capital letter and ends with punctuation, and has a rate between 0.7 and 5 words per second.
We also remove clips without a visible signer, using the detected hands and simple checks for camera cuts and static frames.
The YouTube-ASL segments of setting A (\cref{sec:protocol}) do not use these rules.
As in YouTube-ASL~\cite{youtubeasl2023}, we drop captions that are empty, longer than 300 characters, or shorter than 0.2 or longer than 60 seconds.

\section{Corpus statistics}
\label{app:corpus}
\Cref{tab:corpus_full} extends \cref{tab:corpus} with the domain of each source and the mean duration and number of caption words per clip.

\begin{table*}[!t]
\centering
\small
\caption{Full statistics of the curated training corpus (setting B). Clips and hours are after cleaning and after removing every clip that comes from the same video as a test clip. Duration and words are per clip. The source videos of OpenASL partly overlap with those of YouTube-ASL and YouTube-SL-25, so the total of unique videos is lower than the sum.}
\label{tab:corpus_full}
\begin{tabular}{llrrrrr}
\toprule
Source & Domain & Videos & Clips & Hours & Mean dur.\ (s) & Mean words \\
\midrule
YouTube-ASL~\cite{youtubeasl2023} & web, mixed & 8,790 & 228,399 & 440.4 & 6.9 & 13.4 \\
ASL STEM Wiki~\cite{yin2024stemwiki} & interpreted Wikipedia articles & 229 & 59,109 & 291.7 & 17.8 & 20.8 \\
OpenASL~\cite{shi2022openasl} & web news and vlogs & 1,476 & 70,763 & 169.0 & 8.6 & 14.1 \\
YouTube-SL-25, ASL~\cite{tanzer2025ytsl25} & web, mixed & 4,069 & 83,181 & 152.1 & 6.6 & 13.3 \\
How2Sign~\cite{duarte2021how2sign} & studio, instructional & 2,192 & 30,794 & 58.8 & 6.9 & 17.7 \\
\midrule
Total & & 15,687 unique & 472,246 & 1,112.1 & 8.5 & 14.7 \\
\bottomrule
\end{tabular}
\end{table*}

\section{Test sets}
\label{app:testsets}
\Cref{tab:testsets} lists the test sets of \cref{sec:overlap}.

\begin{table}[!t]
\centering
\small
\caption{Test sets. All are fixed lists shared by every method. The YouTube-ASL split is used in setting B only. In setting B one How2Sign clip without keypoints is skipped (2,342 sentences).}
\label{tab:testsets}
\setlength{\tabcolsep}{4pt}
\resizebox{\linewidth}{!}{%
\begin{tabular}{lrrl}
\toprule
Test set & Sentences & Videos & Split rule \\
\midrule
How2Sign & 2,343 & 149 & Uni-Sign test list \\
ASL STEM Wiki & 4,047 & 25 art. & unseen articles \\
OpenASL & 908 & 414 & official \\
\quad unseen in YouTube-ASL & 300 & 182 & \\
\quad seen in YouTube-ASL & 608 & 232 & \\
YouTube-ASL (ours) & 3,158 & 151 & unseen videos \\
\bottomrule
\end{tabular}}
\end{table}

\section{Sign lexicon}
\label{app:lexicon}
The sign lexicon in the pseudo-gloss prompt (\cref{sec:llmgloss}) merges the glosses of ASL Signbank and the sign labels of ASL Citizen into one entry per English key.
A key is the lowercase gloss with variant markers and trailing digits removed and with hyphens and underscores replaced by spaces, so that variants such as DRINK and DRINKb become one entry.
From Signbank we drop gestures, interjections, depicting signs and name signs, and all pointing entries are merged into three keys, for pointing, possessives and reflexives.
\Cref{tab:lexicon} shows how the two sources overlap, and \cref{tab:lexicon_examples} gives examples of shared and source-only entries.
The 2,731 sign labels of ASL Citizen give 2,305 keys, and 1,670 of them also occur in Signbank, so 635 keys come from ASL Citizen alone.
The overlap is counted on keys, so a sign that is written differently in the two sources counts as two keys, and the true overlap may be somewhat larger.

\begin{table}[!t]
\centering
\small
\caption{Keys of the sign lexicon by source.}
\label{tab:lexicon}
\begin{tabular}{lr}
\toprule
Source & Keys \\
\midrule
ASL Signbank & 2,704 \\
ASL Citizen & 2,305 \\
both & 1,670 \\
ASL Signbank only & 1,034 \\
ASL Citizen only & 635 \\
\midrule
Lexicon & 3,339 \\
\bottomrule
\end{tabular}
\end{table}

\begin{table}[!t]
\centering
\small
\caption{Example entries of the sign lexicon with their original glosses. A dash means the source has no gloss for the key.}
\label{tab:lexicon_examples}
\begin{tabular}{lll}
\toprule
Key & ASL Signbank & ASL Citizen \\
\midrule
\multicolumn{3}{l}{\emph{in both sources}} \\
jump & JUMP & JUMP \\
week & WEEK & WEEK \\
feel & FEEL & FEEL \\
ear & EARix, EARx & EAR \\
\midrule
\multicolumn{3}{l}{\emph{ASL Signbank only}} \\
mistake & MISTAKE & -- \\
sliced bread & SLICED-BREAD & -- \\
take your time & TAKE-YOUR-TIME & -- \\
concern & CONCERN & -- \\
\midrule
\multicolumn{3}{l}{\emph{ASL Citizen only}} \\
march & -- & MARCH \\
yucky & -- & YUCKY \\
frisk & -- & FRISK \\
toilet & -- & TOILET \\
\bottomrule
\end{tabular}
\end{table}

\section{Keypoint extraction}
\label{app:keypoints}
The lightweight whole-body configuration of rtmlib uses a YOLOX-tiny person detector followed by the RTMW-dw-l-m whole-body model~\cite{jiang2024rtmw} at an input size of $256\times192$.
When several people are detected we keep the one with the highest mean keypoint score.
From the 133 whole-body points we keep 9 body points, 21 points per hand and 18 face points, as in Uni-Sign.
Body points are scaled to $[-1,1]$ by the square bounding box of the confident body points, hand points are expressed relative to the wrist and face points relative to the nose, and points with confidence at most 0.3 are set to zero.
Keypoints are available for all but one of the 472,247 training clips.

\section{Hyperparameters of our model}
\label{app:hparams}
\Cref{tab:hparams} lists the hyperparameters of both pretraining stages and of translation.

\begin{table}[h]
\centering
\small
\caption{Hyperparameters of our model. CE denotes the cross-entropy loss and L1 the mean absolute error. All stages use AdamW, gradient clipping at 1.0 and bfloat16.}
\label{tab:hparams}
\setlength{\tabcolsep}{3pt}
\resizebox{\linewidth}{!}{%
\begin{tabular}{lccc}
\toprule
& Stage 1 & Stage 2 & Translation \\
\midrule
Init & random & stage 1 & stage 2 + T5-base \\
Loss & L1 & CE + L1 & CE \\
Mask ratio & 30\% & 15--50\% / 30--50\% & -- \\
Steps & 100K & 80K & 20 epochs \\
Batch & 128 windows & 64 clips & 48 clips \\
Peak learning rate & 1e-4 & 5e-5 & 5e-5 / 1e-4 \\
Warmup & 5K & 5K & 2K \\
Schedule & cosine & cosine & cosine \\
Weight decay & 0.01 & 0.01 & 0.01 \\
Max frames & 256 & 1,024 & 1,024 \\
Parameters & 204M & 235M & 427M \\
\bottomrule
\end{tabular}}
\vspace{2pt}

{\footnotesize The translation learning rate is given for the encoder and for the projection and T5. Setting-A fine-tuning uses 2.5e-5 / 5e-5 with 500 warmup steps.}
\end{table}

\section{Motion check of joint pretraining}
\label{app:motioncheck}
The accuracies in \cref{sec:joint} are measured during joint pretraining, on the validation clips of each setting, before any translation training.
In every clip each pseudo-gloss unit is replaced by [MASK] with probability 0.3, and at least one unit is masked.
All frames stay visible, and the random seed is fixed, so every validation round masks the same units.
The accuracy is the fraction of masked units for which the most probable unit over the vocabulary is the correct one.
We then keep the masked units and give every clip the motion of the next clip in the validation batch, and measure the same accuracy.
If the encoder predicted the masked units from the remaining units alone, this change would not affect the accuracy.
Both numbers are taken from the checkpoint with the best validation accuracy, which is the checkpoint we keep.
Training draws the mask fraction uniformly from $[0.15, 0.5]$, while validation uses a fixed 0.3 so that all rounds are comparable.

\section{Reproduced methods}
\label{app:repro}
\Cref{tab:repro} summarizes the reproduced methods of \cref{sec:repro}.
All of them use the clips and test lists of setting A (\cref{sec:protocol}), and the training length of each method is given below.

\paragraph{YouTube-ASL baseline.}
We follow Uthus~\etal~\cite{youtubeasl2023}.
The input is the 85 keypoints of the paper, 42 hand points, 6 body points and 37 face points, taken from our RTMPose output.
Since RTMPose is 2D, each frame has 170 values instead of the 255 of the paper.
Points with low confidence are set to a constant, and each clip is min-max scaled per axis.
A linear layer feeds the frames into T5-v1.1-Base~\cite{raffel2020t5}.
We train with Adafactor at a constant learning rate of 1e-3 and batch 128 on the YouTube-ASL clips, early-stop at 260K steps when the validation BLEU-4 stops improving, and fine-tune for 20 epochs on each dataset.

\paragraph{Uni-Sign, pose only.}
We train Uni-Sign~\cite{li2025unisign} from scratch with its public code.
Uni-Sign reads the same 69 RTMPose points as our model with the same normalization, encodes each body part with a spatial-temporal graph network, and decodes with mT5-Base~\cite{xue2021mt5}.
Clips longer than 256 frames are subsampled as in the released code.
The pretraining stage of Uni-Sign is run on our YouTube-ASL clips for 20 epochs with batch 512 and learning rate 3e-4, using the translation loss, since the original pretraining script is not released.
The fine-tuning stage then runs for 20 epochs per dataset with batch 32 and learning rate 3e-4, and we report the last epoch.
Label smoothing is 0.2 in both stages, as in the released code.

\paragraph{Uni-Sign, pose and RGB.}
This variant adds the RGB branch of Uni-Sign.
Hand crops of $112\times112$ pixels are cut from the video at the keypoint locations with the released data code, encoded by an ImageNet EfficientNet-B0, and fused with the pose features through the released fusion module.
Starting from the pose-only model after its pretraining stage, we run the second stage of the paper for 5 epochs with batch 128 and learning rate 3e-4 on the same YouTube-ASL clips, and then the same per-dataset fine-tuning as the pose-only model.

\paragraph{SHuBERT.}
SHuBERT~\cite{gueuwou2025shubert} releases its self-supervised encoder but not its translation code.
We use the released encoder and the released DINOv2 feature extractors, build the four input streams (face, left hand, right hand and upper body) with the released crop code, and attach ByT5-Base~\cite{xue2022byt5} through a learned weighted sum over the encoder layers and a linear layer, as described in the paper.
Translation pretraining on the YouTube-ASL clips of setting A follows the paper, with AdamW, peak learning rate 5e-4 (5e-5 for the SHuBERT encoder), weight decay 0.1, label smoothing 0.2, 128 sentences per update, 10K warmup steps and a cosine schedule over 250K steps.
Per-dataset fine-tuning then runs for 20 epochs with learning rate 1e-4 (1e-5 for the encoder).
Because some details are not specified, we report three runs of the same reproduction, A, C1 and C2.
\begin{itemize}
\item \textbf{Run A} locates the face and hand crops with our RTMPose keypoints.
\item \textbf{Run C1} uses the preprocessing of the paper, a person detector and MediaPipe landmarks, so the streams match those the released encoder was trained on.
\item \textbf{Run C2} equals C1, but weight decay is applied only to weight matrices and not to biases and normalization layers, following the common HuggingFace default. The paper states a weight decay of 0.1 without saying which parameters it applies to, and A and C1 apply it to all parameters.
\end{itemize}
Run A was pretrained for 133K steps and fine-tuned from its best validation checkpoint at 110K steps.
C1 and C2 were pretrained for 20 epochs (94,100 steps), the same length as Uni-Sign and our model, and fine-tuned from their best validation checkpoints.
To test more training epochs, we also pretrained C2 for the full 250K steps of the paper (about 53 epochs).

\begin{table*}[h]
\centering
\small
\caption{Reproduced methods under our protocol. All are trained on the same clips and evaluated on the same test lists with beam 4. The last column lists the deviations from the original papers that we are aware of.}
\label{tab:repro}
\setlength{\tabcolsep}{4pt}
\begin{tabular}{p{2.6cm}p{3.6cm}p{2.4cm}p{7.4cm}}
\toprule
Method & Input & Decoder & Deviations from the paper \\
\midrule
YouTube-ASL baseline & 85 RTMPose points, 2D & T5-v1.1-Base & 2D instead of 3D MediaPipe points, beam 4 instead of 5 \\
Uni-Sign, pose & 69 RTMPose points & mT5-Base & pretraining script not released (we use the translation loss), YouTube-ASL only, every second frame with the lightweight RTMPose model \\
Uni-Sign, pose + RGB & + $112^2$ hand crops & mT5-Base & as above \\
SHuBERT (A) & 4 streams located by RTMPose & ByT5-Base & crops located by RTMPose, YouTube-ASL only \\
SHuBERT (C1) & 4 streams, paper preprocessing & ByT5-Base & YouTube-ASL only \\
SHuBERT (C2) & as C1 & ByT5-Base & YouTube-ASL only, weight decay not applied to biases and normalization layers \\
\bottomrule
\end{tabular}
\end{table*}

\section{Validation curves with less translation data}
\label{app:datasize}
\begin{figure}[!t]
\centering
\includegraphics[width=0.85\linewidth]{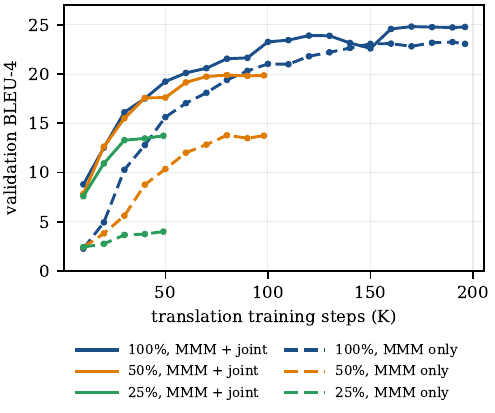}
\caption{Validation BLEU-4 during translation training in setting B, with 100\%, 50\% and 25\% of the training sentences. Solid lines start from the MMM + joint encoder and dashed lines from the MMM-only encoder.}
\label{fig:datasize}
\end{figure}

\Cref{fig:datasize} shows the validation BLEU-4 during translation training for the data-size study of \cref{sec:exp_datasize}.
The jointly pretrained encoder is ahead from the first evaluation, and with less data the MMM-only model learns much more slowly.

\section{Training curves of the reproduced methods}
\label{app:convergence}
\begin{figure*}[!t]
\centering
\includegraphics[width=\textwidth]{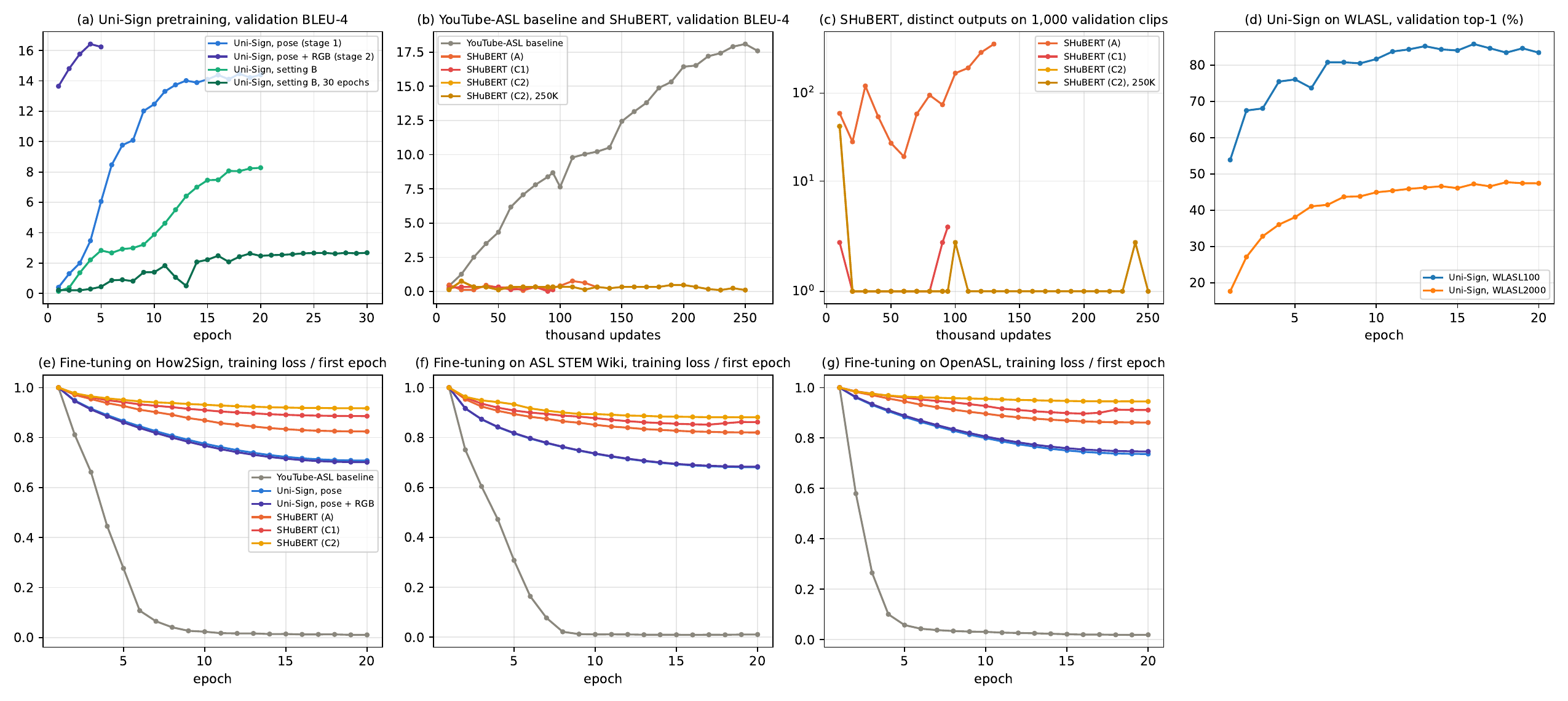}
\caption{Training logs of the reproduced methods. (a) Uni-Sign translation pretraining, validation BLEU-4 per epoch (the pose-only stage 1 and setting B for 20 epochs, setting B also for 30 epochs, the RGB stage 2 for 5 epochs). (b) YouTube-ASL baseline and SHuBERT, validation BLEU-4 against updates (SHuBERT C2 also with a 250K-update schedule). (c) SHuBERT, number of distinct sentences generated for 1,000 validation clips. (d) Uni-Sign on WLASL, validation top-1 accuracy. (e)--(g) Per-dataset fine-tuning, training loss relative to the first epoch.}
\label{fig:repro_curves}
\end{figure*}

A reproduction is only a fair comparison if it is trained to convergence.
\Cref{fig:repro_curves} shows the training logs of every reproduced run in this paper, and \cref{tab:convergence} summarizes them.
Translation pretraining and the WLASL runs are scored on a validation set during training.
Per-dataset fine-tuning follows the protocol of setting A without a validation set, so its evidence is the training loss.

\paragraph{YouTube-ASL baseline.}
Validation BLEU-4 rises to 18.09 at 250K updates, and training stops at 260K updates because it gained less than 0.2 BLEU-4 over the last 20K updates.
During fine-tuning the training loss falls to 1 to 2\% of its first-epoch value on all three datasets, so the model fits its training data completely.
Its low BLEU-4 on How2Sign and STEM is therefore not a lack of training.

\paragraph{Uni-Sign.}
In setting A the pose-only pretraining reaches its best validation BLEU-4 of 14.47 at epoch 18, and the last five epochs stay within 0.4 BLEU-4 of it.
The second stage with RGB peaks at epoch 4 of 5 (16.43, then 16.24).
In all six fine-tuning runs the training loss changes by less than 1.3\% over the last three epochs.
In setting B the validation BLEU-4 is still rising slowly at the end of the 20 epochs (8.06, 8.23 and 8.28 over the last three epochs), as the learning rate decays to zero.
This budget is the same 20 epochs over the same data that our model uses.
To test more training epochs, we also trained setting B for 30 epochs with otherwise the same recipe.
It does not do better.
Its validation BLEU-4 stays between 2.4 and 2.7 from epoch 19 on and ends at 2.68.
Its test BLEU-4 is lower on every test set, for example 0.17 against 3.43 on STEM and 1.29 against 3.35 on How2Sign.
We therefore report the 20-epoch run.
On WLASL the validation accuracy levels off after about 12 epochs, at 83 to 86\% on WLASL100 and 46 to 48\% on WLASL2000.

\paragraph{SHuBERT.}
For the test sets, C1 also produces the single sentence ``That is all for today.'' for all 908 OpenASL clips, and C2 produces 9 distinct sentences there.
With teacher forcing, the loss of run C1 is the same with its own input, with the input of another clip and with an all-zero input (1.269 in all three cases).
We found that weight decay on all parameters shrinks the normalization layers of ByT5 to a small fraction of their pretrained values, which motivated run C2.
C2 avoids this but still collapses, so weight decay is not the only cause.
C2 pretrained for the full 250K steps of the paper does not recover either, and its validation BLEU-4 peaks at 0.75 after 20K steps and ends at 0.10.
Validation BLEU-4 stays below 0.8 for all three runs.
From 20K updates on, SHuBERT runs C1 and C2 produce one to three distinct sentences for the 1,000 validation clips, and C2 trained with a 250K-update schedule still produces one or two distinct sentences up to the end.
Its training loss is still falling at the end of this run, yet its validation BLEU-4 ends at 0.10.
SHuBERT run A produces more distinct outputs late in training, up to 363, but its validation BLEU-4 falls at the same time, from 0.76 at 110K updates to 0.31 at 130K updates.
During fine-tuning the training loss falls by only 5 to 18\%, against 25 to 32\% for Uni-Sign.
Longer training does not change the outcome, so the collapse of SHuBERT is not a lack of training.

\begin{table*}[!t]
\centering
\small
\caption{Training summary of every reproduced run. H2S, STEM and OA denote How2Sign, ASL STEM Wiki and OpenASL. Validation BLEU-4 for translation pretraining and top-1 accuracy (\%) for WLASL. Loss change compares the mean training loss of the last 15\% of the run with the 15\% before it.}
\label{tab:convergence}
\setlength{\tabcolsep}{4pt}
\resizebox{\textwidth}{!}{%
\begin{tabular}{llcccc}
\toprule
Method & Run & Length & Best validation (at) & Last validation & Loss change \\
\midrule
YouTube-ASL baseline & pretraining & 260K updates, early stop & 18.09 (250K) & 17.59 & \\
& fine-tuning H2S / STEM / OA & 20 epochs & -- & -- & loss at 1--2\% of start \\
Uni-Sign, pose & stage 1 & 20 epochs & 14.47 (ep.\ 18) & 14.41 & $-$1.0\% \\
& fine-tuning H2S / STEM / OA & 20 epochs & -- & -- & $-$1.2 / $-$1.0 / $-$1.1\% \\
Uni-Sign, pose + RGB & stage 2 & 5 epochs & 16.43 (ep.\ 4) & 16.24 & \\
& fine-tuning H2S / STEM / OA & 20 epochs & -- & -- & $-$1.2 / $-$1.0 / $-$1.0\% \\
Uni-Sign, setting B & all data & 20 epochs & 8.28 (ep.\ 20) & 8.28 & $-$0.8\% \\
Uni-Sign, setting B & all data, more epochs & 30 epochs & 2.68 (ep.\ 30) & 2.68 & $-$0.5\% \\
Uni-Sign, WLASL100 & fine-tuning & 20 epochs & 85.80 (ep.\ 16) & 83.43 & $-$0.3\% \\
Uni-Sign, WLASL2000 & fine-tuning & 20 epochs & 47.70 (ep.\ 18) & 47.40 & $-$0.7\% \\
SHuBERT (A) & pretraining & 133K updates & 0.76 (110K) & 0.31 & $-$1.6\% \\
SHuBERT (C1) & pretraining & 94K updates & 0.33 (40K) & 0.12 & $-$0.4\% \\
SHuBERT (C2) & pretraining & 94K updates & 0.75 (20K) & 0.33 & $-$0.5\% \\
SHuBERT (C2) & pretraining, more epochs & 250K updates & 0.75 (20K) & 0.10 & $-$5.8\% \\
\bottomrule
\end{tabular}}
\end{table*}

\end{document}